\documentclass{article}

\usepackage[final]{corl_2020} 
\usepackage{graphicx}
\usepackage{comment}
\usepackage{amsmath,amssymb} 
\usepackage{color}

\usepackage[all]{xy}
\usepackage[labelfont=bf]{caption}
\usepackage{subcaption}

\usepackage{booktabs}
\usepackage{multirow}
\usepackage{multicol}

\usepackage{floatrow}
\usepackage{placeins}
\usepackage{etoc}

\usepackage{enumitem}

\title{SAM: Squeeze-and-Mimic Networks for Conditional Visual Driving Policy Learning} 

\author{
  Albert Zhao$^{*1}$ \hspace{13mm} Tong He$^{*1}$ \hspace{11mm} Yitao Liang$^{1}$ \\ \textbf{Haibin Huang$^{2}$ \hspace{4mm} Guy Van den Broeck$^{1}$ \hspace{2mm} Stefano Soatto$^{1}$}\\
  $^{1}$University of California, Los Angeles, $^{2}$Kuaishou Technology\\
  \texttt{\{azzhao,simpleig,yliang,guyvdb,soatto\}@cs.ucla.edu,}\\
  \texttt{jackiehuanghaibin@gmail.com}\\
  \footnotesize{* equal contribution}
}

\begin{document}
\etocdepthtag.toc{mtchapter}
\maketitle

\begin{abstract}
We describe a policy learning approach to map visual inputs to driving controls conditioned on turning command that leverages side tasks on semantics and object affordances via a learned representation trained for driving. To learn this representation, we train a squeeze network to drive using annotations for the side task as input. This representation encodes the driving-relevant information associated with the side task while ideally throwing out side task-relevant but driving-irrelevant nuisances. We then train a mimic network to drive using only images as input and use the squeeze network's latent representation to supervise the mimic network via a mimicking loss. Notably, we do not aim to achieve the side task nor to learn features for it; instead, we aim to learn, via the mimicking loss, a representation of the side task annotations directly useful for driving. We test our approach using the CARLA simulator. In addition, we introduce a more challenging but realistic evaluation protocol that considers a run that reaches the destination successful only if it does not violate common traffic rules.
\end{abstract}

\section{Introduction}

\label{sec:intro}

Driving is a complex endeavor that consists of many sub-tasks such as lane following, making turns, and stopping for obstacles. A traditional strategy to tackle this complexity is to split the driving policy into two stages: perception (i.e. estimating a manually chosen representation of the scene) and control (i.e. outputting low-level controls using hand-coded rules). Though this approach is interpretable with easy-to-diagnose failures, it suffers from various issues: the representations may be suboptimal and difficult to estimate while hand-coded rules may struggle to handle the full range of driving scenarios \cite{Shwartz2016saferl,liang2018cirl,Sauer2018ConditionalAL}. These drawbacks motivate academia to explore learning approaches for driving as they are not restricted by hand-coded decisions. 
A major challenge for learning approaches is obtaining a useful representation. A simple approach is to directly learn a driving model that maps from image to low-level control. Though this approach is easy to implement, it suffers from poor generalization due to overfitting to nuisances \cite{Li2018RethinkingSM,Codevilla_2019_ICCV}.
\begin{figure}[t]
 \centering
 \begin{subfigure}[b]{0.24\textwidth}
    \centering
    \entrymodifiers={+[o][F-]}
    \centerline{\xymatrix@R-0.5pc@C-0.75pc{
      I\ar[r]  & S \ar[r] & A}}
    \caption{Two-stage}
    \label{fig:two-stage}
 \end{subfigure} 
 \begin{subfigure}[b]{0.24\textwidth}
    \centering
    \centerline{\xymatrix@R-0.5pc@C-0.75pc{
               &                 & *+[o][F-]{S} \\
      *+[o][F-]{I}\ar[r]  & *+[o][F-]{Z} \ar[r]\ar@{-->}[ur] & *+[o][F-]{A}}}
     \caption{Multi-task}
     \label{fig:multi-task} 
  \end{subfigure}
  \begin{subfigure}[b]{0.24\textwidth}
    \centering
    \centerline{\xymatrix@R-0.5pc@C-0.75pc{
      *+[o][F-]{I}\ar@{-->}[r]  & *+[o][F-]{Z_I^S} \ar@{-->}[r]       & *+[o][F-]{S} \\
      *+[o][F-]{I}\ar[r]  & *+[o][F-]{Z} \ar[r]\ar@{-->}[u] & *+[o][F-]{A}}}
     \caption{Feature mimic}
     \label{fig:feature-mimick}
  \end{subfigure}
  \begin{subfigure}[b]{0.24\textwidth}
    \centering
    \centerline{\xymatrix@R-0.5pc@C-0.75pc{
      *+[o][F-]{S}\ar@{-->}[r]  & *+[o][F-]{Z_S^A} \ar@{-->}[r]       & *+[o][F-]{A} \\
      *+[o][F-]{I}\ar[r]  & *+[o][F-]{Z} \ar[r]\ar@{-->}[u] & *+[o][F-]{A}}}
     \caption{Ours}
     \label{fig:ours}
  \end{subfigure}
  \vspace{5pt}
  \caption{Approaches for leveraging side tasks for driving. $I, S,$ and $A$ represent the image, side task annotations, and low-level controls; $Z$, $Z_I^S$, and $Z_S^A$, latent representation, mimicked features, and the squeezed encoding of the side task annotations. Dashed arrows, depending on direction, represent branches or supervision used only during training. Unlike other methods that aim to achieve the side task (two-stage, multi-task) or learn features for it (feature mimic), our method, SAM, learns only driving-relevant context from the side task annotations by directly feeding them as input during training. Also, note that SAM does not suffer from side task estimation errors encountered by other methods because we never explicitly estimate any side task annotations during training/inference.
  }
\label{fig:diffstrategies}
\vspace{-10pt}
\end{figure}
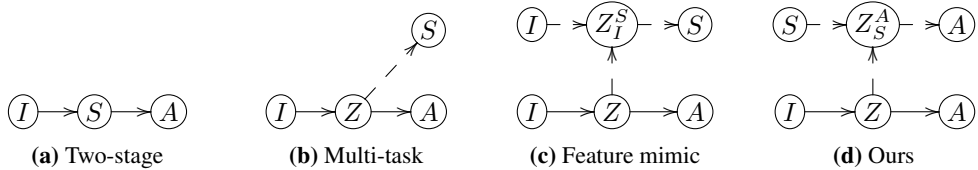

To improve generalization, various approaches (Fig. \ref{fig:diffstrategies}) propose to leverage complex side tasks such as semantic segmentation which provide driving-relevant contextual knowledge in an easily accessible form (i.e. without photometric nuisances). The most straightforward of these approaches is the two-stage approach \cite{Muller2018DPT,huang2019learning,behl2020label,Bansal2019ChaffeurNet} (Fig. \ref{fig:two-stage}), which learns to achieve the side task and then uses the output from the side task to estimate controls. However, this method suffers from errors in directly estimating the complex side task, leading to unrecoverable failures in the driving policy.

To avoid these estimation errors at test time, multi-task learning approaches \cite{Xu2016EndtoEndLO,Li2018RethinkingSM} (Fig. \ref{fig:multi-task}) do not estimate the side task at test time but instead incorporate driving-relevant context by training a shared representation to achieve both the side task and the main driving task. However, they suffer from the fact that the side task is distinct from the main task (i.e. side task-main task mismatch), and hence, there exists information relevant to the side task but not to the main task. For example, the side task of semantic segmentation encourages the learned representation to be informative for every pixel's class label, but knowing such pixel-wise information is unnecessary for driving. Instead, what is more important is knowing that an obstacle of a certain class is in front of the agent. Hence, as multi-task approaches aim to achieve both the side task and the driving task, they encourage the representation to contain side task-relevant but driving-irrelevant nuisances, hurting performance.

Feature mimicking \cite{Hou2019FeatureMimick} (Fig. \ref{fig:feature-mimick}) incorporates contextual information while reducing the reliance on ground-truth annotations for the side tasks (side task annotations) by proposing instead to mimic features of a network (pre)trained to achieve the side task. However, this method suffers from a similar issue (side task-main task mismatch) to multi-task methods: as feature mimicking encourages the representation, via the mimicked features, to be informative for achieving the side task, it encourages the learning of side task relevant but driving-irrelevant nuisances. Moreover, all these previous methods (Fig. \ref{fig:diffstrategies}) inevitably suffer from side-task estimation errors during training/inference as they rely on training models to explicitly estimate the side task.

In contrast with previous approaches, we propose a method (Fig. \ref{fig:ours}) that effectively leverages the side task for conditional driving policy learning without trying to achieve the side task or learn features trained to do so. Instead, we aim to learn a representation of the side task annotations that is directly useful for driving. We first train a squeeze network to drive using ground-truth side task annotations as input and extract its deep features. As these features are trained only for driving and not to achieve the side task, they contain only driving-relevant knowledge and not the side task-relevant but driving-irrelevant nuisances. However, we cannot use the squeeze network to drive as ground-truth side task annotations are unavailable at test time. To train a driving policy that does not require these annotations while still leveraging the side task, we train a mimic network to drive using only image and self-speed as input while pushing it to learn the squeeze network's embedding via a mimicking loss. Hence, the mimicking loss encourages the mimic network's representation to learn only the driving-relevant context associated with the side task. Our overall approach, squeeze-and-mimic networks (\emph{SAM}), leverages the side task in an effective way to train a conditional driving policy without achieving the side task or learning features for it, avoiding the pitfalls of previous methods such as estimation errors and learning side task relevant but driving-irrelevant nuisances.

We note that for all the aforementioned methods (Fig.~\ref{fig:diffstrategies}), the side tasks should not discard important driving-relevant information contained in the image input. Indeed, they should contain this information in an easily accessible form (i.e. without photometric nuisances) so that the intermediate representation can easily learn this information. Two important yet complementary types of information are object class, useful for tasks like lane following, and braking behavior. Hence, for our choice of side task annotations, we select semantic segmentation and {\em stop intention values,} which indicate whether to stop for different types of obstacles such as pedestrians, traffic lights, or cars.

We test our agent on online benchmarks using the photorealistic CARLA simulator \cite{Dosovitskiy2017Carla}. Furthermore, we find previous benchmarks lacking as we do not think one should be rewarded for reaching the destination while driving through red lights or into the opposite lanes. Therefore we introduce a more realistic evaluation protocol that rewards successful routes only if they do not violate basic traffic rules. To summarize, our main contributions are as follows: 
\begin{itemize}[leftmargin=15pt]
   \item a method for conditional driving policy learning that leverages a side task and learns only driving-relevant knowledge from its annotations, avoiding the pitfalls of existing methods that aim to achieve the side task or learn features for it.
   \item the combination of complementary side tasks: semantic segmentation, which provides basic class concepts for subtasks like lane following, and stop intentions, which provide causal information linking braking to hazardous driving scenarios. While both tasks lead to improvement over the baseline, we show that the best performance is achieved only when they are used jointly.
   \item an evaluation protocol, Traffic-school, that fixes the flaws of prior benchmarks. Though CARLA reports various driving infractions, these statistics are scattered and hard to analyze. Thus, previous benchmarks usually compute the route success rate, but they ignore several risky driving behaviors. In contrast, our protocol sets a high standard for the autonomous agent, penalizing previously ignored infractions such as violating red lights and running into the sidewalk.
\end{itemize}
Code is available at \url{https://github.com/twsq/sam-driving}.

\section{Related Work}

Autonomous driving has drawn significant attention for several decades \cite{Pomerleau1989alvinn,LeCun2005OffRoadOA,Silver2010LearningFDAN}. Generally, there exist three different approaches to driving: modular, direct perception, and end-to-end learning.

\emph{Modular pipelines} form the most popular approach \cite{Ullman1980ADP,Paden2016Survey,FrankeADBook,dickmanns1989dynamic}, separating driving into two components: perception modules \cite{Geiger2013IJRR,he2019mono3d++,he2019geonet} that estimate driving-related side tasks and control modules. Recently, \cite{Muller2018DPT,huang2019learning,behl2020label,Bansal2019ChaffeurNet} use semantic segmentation and environment maps as side task annotations in a modular pipeline. However, these annotations are complex and high-dimensional. To simplify the annotations and alleviate the annotation burden, a recent work LEVA \cite{behl2020label} proposes to use coarse segmentation masks. The mentioned modular approaches have the advantage of outputting interpretable representations in the form of estimated side task annotations (perception outputs), allowing for the diagnosis of failures. However, they suffer from perception and downstream control errors, \textit{and} use manually chosen representations that are potentially suboptimal for driving. 

\emph{Direct perception} \cite{Gibson1979affordances} methods similarly separate the driving model into two parts, but unlike modular approaches, they avoid complex representations, opting for compact intermediate representations instead. \cite{Chen2015DeepDriving} and \cite{Sauer2018ConditionalAL} (CAL) both train a network to estimate affordances (distance to vehicle,  center-of-lane, etc.) linked to controls, for car racing and urban driving, respectively. Similar to modular approaches, the intermediate representation is hand-crafted, with no guarantees that it is optimal. 

\emph{End-to-end methods} \cite{Pomerleau1989alvinn,LeCun2005OffRoadOA,Silver2010LearningFDAN,Bojarski2016nvidia} learn a direct mapping from image to control and are most related to our work. CIL \cite{codevilla2018end} builds upon offline behavioral cloning, solving the ambiguity problem at traffic intersections by conditioning on turning commands. \cite{Codevilla_2019_ICCV} analyzes issues within the CIL approach and proposes an improved version, CILRS. However, both methods fail to generalize to dense traffic and suffer from the covariate shift between offline training data and closed-loop evaluation. To reduce this covariate shift, various methods collect data online using DAgger \cite{Ross2011Dagger} imitation learning \cite{Chen2019lbc,prakash2020exploring} or reinforcement learning (RL) \cite{Pan2017VTRRL,liang2018cirl,toromanoff2019end,ohn2020learning}. Recently, \cite{toromanoff2019end} combines RL with a feature extractor pretrained on affordance estimation while LSD \cite{ohn2020learning} combines RL with a multimodal driving agent trained via mixture of experts. However, the above methods all use online training, which is expensive and unsafe and can be performed only in simulation \cite{Dosovitskiy2017Carla}. Due to these drawbacks, we focus on offline methods as offline and online methods are not directly comparable.

Previous methods have also leveraged side task annotations to improve generalization and acquire context knowledge. As discussed in the Introduction and shown in Fig. \ref{fig:diffstrategies}, the two-stage, multi-task and feature mimic methods \cite{behl2020label,Xu2016EndtoEndLO,Li2018RethinkingSM,Hou2019FeatureMimick} all suffer from side task estimation errors and tend to learn driving-irrelevant nuisances (side task-main task mismatch).
We overcome these issues by never aiming to achieve the side task; instead, we mimic a representation squeezed from side task annotations and trained for driving. Similar to our method, LBC \cite{Chen2019lbc} also mimics a driving agent. However, this agent, unlike ours, takes expensive 3D side task annotations as input. Moreover, LBC and SAM use different methods of mimicking: LBC mimics the agent's \emph{controls} while SAM mimics the agent's \emph{embeddings}. Hence, our method directly encourages the model's representation to contain driving-related context associated with the side task while LBC does not. To illustrate the benefits of our mimicking method, we also test SAM using LBC's control mimicking.

\section{Method}

\label{sec:method}

\begin{figure*}[t]
   \begin{center}
   \includegraphics[width=0.95\textwidth]{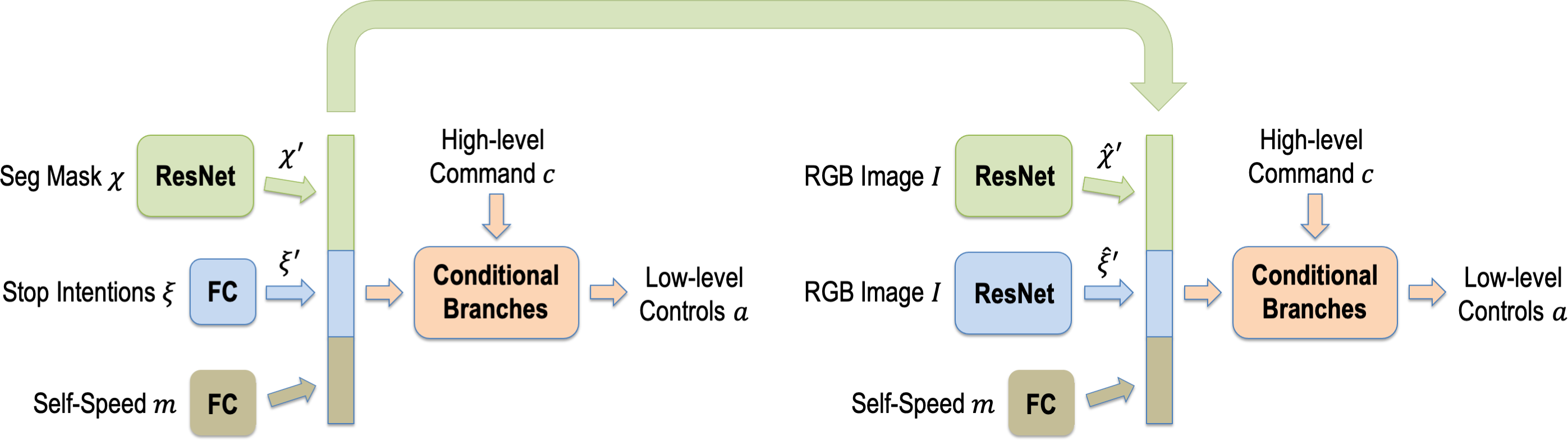}
   \end{center}
   \vspace{-8pt}
   \caption{Architecture overview of our squeeze and mimic networks. Both use a three-branch structure. (Left) Squeeze network: processes semantic segmentation, stop intention values, and self-speed in separate branches and feeds their concatenated representation to a command-conditioned driving module. (Right) Mimic network: has a similar structure but takes in only RGB image and self-speed. (Arrow) $l_2$ loss is enforced between the segmentation and intentions embeddings of the squeeze network and those of the ResNet branches, which each take in image, of the mimic network.}
\label{fig:pipeline}
\end{figure*}

The overall pipeline of our method {\em SAM} (squeeze-and-mimic networks) is shown in Fig. \ref{fig:pipeline}. The goal of conditional driving policy learning is to learn a policy $(I, m, c) \mapsto a$ that outputs low-level controls $a$ given image $I$, self-speed $m$, and turning command $c = $ $\{{\rm follow, left, right, straight}\}$. Low-level continuous controls $a = (b, g, s)$ consist of brake $b$, gas $g$, and steering angle $s$.

For our method, we additionally leverage side task annotations $S = (\chi, \xi$), in particular semantic segmentation $\chi$ and stop intention values $\xi$ provided by the CARLA simulator, to train a conditional driving policy. The stop intentions $\xi = (v, p, l) \in \left [0, 1 \right ]^{3}$ indicate how urgent it is for the agent to brake in order to avoid hazardous traffic situations such as collision with vehicles, collision with pedestrians, and red light violations, respectively. They are like instructions given by a traffic-school instructor, which inform you of the causal relationships between braking and different types of dangerous driving scenarios, complementing the semantic segmentation $\chi$, which contains concepts of object class identity for tasks like lane following. Hence, the training dataset for our method consists of temporal tuples $\{I_{t},m_{t},c_{t},\chi_{t},\xi_{t}, a_t\}_{t=1}^{T}$, collected from a rule-based autonomous driving agent with access to all internal state of the CARLA driving simulator.

In our method {\em SAM}, we first squeeze the side task annotations $S_t$, consisting of segmentation masks $\chi_t$ and three-category stop intentions $\xi_{t} = (v_t, p_t, l_t)$, into a latent representation ($\chi_{t}^{'},\xi_{t}^{'}$) containing driving-relevant info and not driving-irrelevant nuisances by training a squeeze network to drive using side task annotations, self-speed, and turning command as input: $(S_{t}, m_{t}, c_{t}) \mapsto a_{t}$. We then train a mimic network $(I_{t}, m_{t}, c_{t}) \mapsto a_{t}$ to drive with no access to side task annotations while encouraging this network's embedding ($\hat{\chi}_{t}^{'},\hat{\xi}_{t}^{'}$) to mimic the squeeze network's embedding. Notably, the squeeze network does not take image as input; this is so that its latent representation, which is used to supervise the mimic network, does not contain photometric nuisances from the image.

\subsection{Squeeze Network}

The task of our squeeze network is to squeeze the ground-truth segmentation masks $\chi_{t}$ and three-category stop intentions $\xi_{t} = (v_{t},p_{t},l_{t})$ into a representation that contains contextual driving information but with nuisances removed. Such a representation will later be used to supervise the mimic network via a mimicking loss. As shown in Fig.~\ref{fig:pipeline}, the squeeze network uses a three-branch  architecture for estimating controls $a_{t} = (b_{t},g_{t},s_{t})$. The first branch uses a ResNet34 backbone to squeeze the segmentation mask input, which provides concepts of object class useful for tasks like lane following. We use the middle fully-connected (FC) branch to process the three-category stop intentions, which complement the segmentation mask by informing the agent of the causal relationships between braking behaviors and the presence of objects in the context of safe driving. The lower FC branch ingests self-speed, which provides context for speeding up / slowing down \cite{Codevilla_2019_ICCV}.

The latent feature vectors from the three branches are concatenated and fed into a driving module, composed of several FC layers and a conditional switch \cite{codevilla2018end} that chooses one out of four different output branches depending on the given turning signals $c_{t}$. The four output branches share the same network architecture but with separately learned weights. We use $\ell_1$ losses for training:
\begin{equation}
L_{control} = \lambda_{1}|\hat{b}_t-b_t| + \lambda_{2}|\hat{g}_t-g_t| + \lambda_{3}|\hat{s}_t-s_t|
\label{control_loss}
\end{equation}
\noindent where $(\hat{b}_{t},\hat{g}_{t},\hat{s}_{t})$ are estimated controls of brake, throttle, steering angle respectively,  and $(b_{t},g_{t},s_{t})$ are ground-truth controls. $\lambda_i$'s are weights for loss terms.

\subsection{Mimic Network}

The mimic network does not have direct access to side task annotations, but instead observes a single RGB image $I_{t}$, self-speed measurement $m_{t}$, and high-level turning command $c_{t}$ and mimics the squeeze network's latent embedding to learn driving-relevant context without learning driving-irrelevant nuisances. Its goal is also to estimate low-level controls $a_{t}$, for which it also adopts a three-branch network. The first and the second branches are now both ResNet34 backbones, pretrained on ImageNet, that separately take in image $I_{t}$ and output latent embeddings $({\hat{\chi}_{t}^{'}},{\hat{\xi}_{t}^{'}})$. We adopt separate branches as this separation aids mimicking the embeddings of the segmentation masks, which are pixel-wise and provide object class identity, and of the intention values, which are global and provide causal links between braking and dangerous situations, respectively. 

When training the mimic network, as illustrated in Fig.~\ref{fig:pipeline}, besides applying the $\ell_1$ control loss  (Eq.~\eqref{control_loss}), we enforce $\ell_2$ mimicking losses with weights $\lambda_{i}^{'}$ to encourage the mimic network's embeddings $({\hat{\chi}_{t}^{'}},{\hat{\xi}_{t}^{'}})$ to mimic the squeeze network's embeddings  $({\chi_{t}^{'}}, {\xi_{t}^{'}})$ of the side task annotations:
\begin{equation}
L_{mimic} = \lambda_1^{'}\left \| \hat{\chi}_{t}^{'} -{\chi_{t}^{'}} \right \|_2^2 + \lambda_2^{'}\left \| \hat{\xi}_{t}^{'} - {\xi_{t}^{'}} \right \|_2^2.
\label{latent_loss}
\end{equation}
The mimicking and control losses are combined to learn a driving policy:
$L = L_{control} + L_{mimic}.$ 

\subsection{Implementation Details}

We implement our approach using CARLA 0.8.4 \cite{Dosovitskiy2017Carla}. To train both the squeeze and mimic networks, we use a batch size of 120 and the ADAM optimizer with initial learning rate 2e-4. We use an adaptive learning rate schedule, which reduces the learning rate by a factor of 10 if the training loss has not decreased in 1000 iterations. We use a validation set for early stopping, validating every 20K iterations and stopping training when the validation loss stops decreasing.

Regarding time complexity, our mimic network demonstrates real-time performance (59 FPS) on a Nvidia GTX 1080Ti. Training the squeeze network takes 10 hours on a Nvidia GTX 1080Ti while the mimic network trains in 1 day on a Nvidia Titan Xp.

\section{Experiments}

\label{sec:experiments}

We demonstrate the effectiveness and generalization ability of our method by evaluating it on standard closed-loop evaluation benchmarks. In section 4.1, we compare our model against a suite of competing approaches. Concerned about the fact that multiple driving infractions are not penalized in existing benchmarks, we then introduce a more realistic new evaluation standard {\em Traffic-school} in section 4.2. In addition, in sections 4.3 and 4.4, we conduct ablation studies on the strategy for leveraging the side task and individual types of side task annotations. For all benchmarks, we evaluate in four combinations of towns, which differ in road layout and visual nuisances, and weathers to test generalization: training conditions, new weather, new town, and new town/weather.

\subsection{Comparison with State-of-the-Art on CARLA}

We first test our method on the NoCrash and Traffic-light benchmarks~\cite{Codevilla_2019_ICCV}. We additionally present results on the saturated old CARLA~\cite{codevilla2018end} benchmark in Supp. Mat. For NoCrash, we also include, for completeness, results for LSD~\cite{ohn2020learning}, LBC~\cite{Chen2019lbc}, and LEVA~\cite{behl2020label} even though they are not directly comparable to our method. LSD~\cite{ohn2020learning} uses online training, which is unsafe, and hence, is dependent on a driving simulator, while our method uses only offline training, avoiding these drawbacks. LBC~\cite{Chen2019lbc} also uses online training and in addition uses expensive ground-truth 3D maps, collecting which requires “accessing the ground-truth state of the environment,” which “is difficult in the physical world” \cite{Chen2019lbc}. In contrast, our method uses only 2D segmentation and stop intentions. Finally, LEVA uses different pretraining (using MS-COCO segmentation dataset) and a significantly larger architecture ($\sim$100M parameters for LEVA vs. $\sim$45M for our model). For fair comparisons, in Tab.~\ref{nocrash}, we also test LEVA (see ``efficient seg") and LBC (see ``control mimic") using comparable backbones and the same dataset as SAM. Details of these baselines are in Supp. Mat.

\begin{table*}[t!]
\centering
\vspace{-6pt}
\caption{Results on NoCrash~\cite{Codevilla_2019_ICCV} benchmark. Empty, regular and dense refer to three levels of traffic. We show navigation success rate $(\%)$ in different test conditions. Due to simulator randomness, all methods are evaluated 3 times. Bold indicates best among comparable methods. Italics indicate best among all methods, including those whose results are provided solely for completeness such as LSD~\cite{ohn2020learning}, LEVA~\cite{behl2020label}, and  LBC~\cite{Chen2019lbc}. Note that LSD~\cite{ohn2020learning} and LEVA~\cite{behl2020label} report results only on the new town. We also include comparable baselines (marked with *), efficient seg and control mimic, for LEVA~\cite{behl2020label} and LBC~\cite{Chen2019lbc}. The average error for {\em our model SAM} is $\textbf{28\%}$, which is $\textbf{30\%}$ better than {\em the next-best CILRS}, $\textbf{40\%}$, in terms of relative failure reduction.}
\resizebox{1.0 \textwidth}{!}{
\begin{tabular}{lccclccclccclccclc}
\toprule
                & \multicolumn{3}{c}{\ Training} & \multicolumn{3}{c}{\ New weather} & \multicolumn{3}{c}{\ New town} & \multicolumn{3}{c}{\ New town/weather} &  \multirow{2}{*}{\ Mean}     \\
Method          & Empty       & Regular      & Dense      & \ \ Empty       & Regular      & Dense      & \ \ Empty       & Regular      & Dense      &  \ \ Empty       & Regular      & Dense      & \ \  \\ 
\midrule
\textbf{Comparable Baselines} \\
CIL~\cite{codevilla2018end}   & $79\pm1$         & $60\pm1$          & $21\pm2$        &  \ \ $83\pm2$         & $55\pm5$          & $13\pm4$        & \ \ $48\pm3$         & \ $27\pm1$          & $10\pm2$        &  \ \ $24\pm1$         & $13\pm2$          & \  $2\pm0$        & \ \ $36\pm0.7$  \\
CAL~\cite{Sauer2018ConditionalAL}   & $81\pm1$         & $73\pm2$          & $42\pm3$        & \ \ $85\pm2$         & $68\pm5$          & $33\pm2$        & \ \ $36\pm6$         & \ $26\pm2$          & \ $9\pm1$        & \ \ $25\pm3$         & $14\pm2$          & $10\pm0$        & \ \ $42\pm0.8$  \\
MT~\cite{Li2018RethinkingSM}    & $84\pm1$         & $54\pm2$          & $13\pm4$        &   \ \ $58\pm2$         & $40\pm6$          & \ $7\pm2$        & \ \ $41\pm3$         & \ $22\pm0$          & \ $7\pm1$        & \ \ $57\pm0$         & $32\pm2$          & $14\pm2$        & \ \ $36\pm0.8$  \\
Efficient seg*~\cite{behl2020label} & \textbf{100} $\pm$ \textbf{0} & 82 $\pm$ 1 & 38 $\pm$ 5 & \ \ 80 $\pm$ 2 & 72 $\pm$ 3 & 23 $\pm$ 3 & \ \ 91 $\pm$ 1         & \ 58 $\pm$ 3         & 15 $\pm$ 4       &  \ \ 68 $\pm$ 3       & 47 $\pm$ 5         & 19 $\pm$ 6        & \ \ 58 $\pm$ 1.0           \\
Control mimic*~\cite{Chen2019lbc} & \textbf{100} $\pm$ \textbf{0} & 84 $\pm$ 1 & 31 $\pm$ 3 &  \ \ 99 $\pm$ 1 & 76 $\pm$ 6 & 27 $\pm$ 6 & \ \ 84 $\pm$ 1 & \ 47 $\pm$ 2 & 11 $\pm$ 3 & \ \ \textbf{83} $\pm$ \textbf{3} & 51 $\pm$ 1 & 10 $\pm$ 2 & \ \ $59\pm0.9$ \\
CILRS~\cite{Codevilla_2019_ICCV} & $97\pm2$         & $83\pm0$          & $42\pm2$        &  \ \ $96\pm1$         & $77\pm1$          & $39\pm5$        & \ \ $66\pm2$         & \ $49\pm5$          & $23\pm1$        & \ \ $66\pm2$         & $56\pm2$          & $24\pm8$        &  \ \ $60\pm1.0$  \\
SAM             & \textbf{100} $\pm$ \textbf{0}         & \textbf{94} $\pm$ \textbf{2}          & \textbf{54} $\pm$ \textbf{3}        & \ \textbf{100} $\pm$ \textbf{0}         & \textbf{89} $\pm$ \textbf{3}          & \textbf{47} $\pm$ \textbf{5}        & \ \ \textbf{92} $\pm$ \textbf{1}         & \ \textbf{74} $\pm$ \textbf{2}          & \textbf{29} $\pm$ \textbf{3}        & \ \ \textbf{83} $\pm$ \textbf{1}         & \textbf{68} $\pm$ \textbf{7}          & \textbf{29} $\pm$ \textbf{2}        & \ \ \textbf{72} $\pm$ \textbf{0.9}  \\ 
\textbf{Listed for Completeness} \\
LSD~\cite{ohn2020learning} &  N/A         & N/A          & N/A        & \ \ \ \ N/A         & N/A          & \ N/A        & \ \ $94\pm1$         & \ $68\pm2$          & $30\pm4$        & \ \ $95\pm1$         & $65\pm4$          & $32\pm3$        & \ \ N/A  \\
LEVA~\cite{behl2020label} & N/A         & N/A          & N/A        & \ \ \ \ N/A         & N/A          & \ N/A        &  \ \ $87\pm1$         & \ $82\pm1$          & $41\pm1$        & \ \ $79\pm1$         & $71\pm1$          & $32\pm5$        & \ \ N/A  \\ 
LBC~\cite{Chen2019lbc} & \textit{100} $\pm$ \textit{0}         & \textit{99} $\pm$ \textit{1}          & \textit{95} $\pm$ \textit{2}        & \ \textit{100} $\pm$ \textit{0}         & \textit{99} $\pm$ \textit{1}          & \textit{97} $\pm$ \textit{2}        &  \ \textit{100} $\pm$ \textit{0}         & \ \textit{96} $\pm$ \textit{5}          & \textit{89} $\pm$ \textit{1}        & \ \textit{100} $\pm$ \textit{2}         & \textit{94} $\pm$ \textit{4}          & \textit{85} $\pm$ \textit{1}        & \ \ \textit{96} $\pm$ \textit{0.6}  \\
\bottomrule
\end{tabular}}
\vspace{-2mm}
\label{nocrash}
\end{table*}

\noindent \textbf{NoCrash} \; We report results on the NoCrash benchmark in Tab.~\ref{nocrash}, where the metric is navigation success rate in three different levels of traffic: empty, regular, and dense. A route is considered successful for NoCrash only if the agent reaches the destination within the time window without crashing. Though the {\em second-best CILRS} outperforms the other comparable baselines, our method, with average error $\textbf{28\%}$, still achieves a relative failure rate reduction of $\textbf{30\%}$ over CILRS, which has average error $\textbf{40\%}$. In particular, our method achieves especially large performance gains over CILRS in new town and under regular and dense traffic. These gains demonstrate the effectiveness of both our choice of side task annotations (semantic segmentation for basic class concepts and stop intentions to aid braking) and our method of leveraging them. We also note that our method outperforms MT~\cite{Li2018RethinkingSM} (a multi-task method), control mimic, and efficient seg. This comparison illustrates the advantages of our method over other methods of using the side task annotations; our method learns only driving-relevant context associated with the side task. On the other hand, multi-task learning, efficient seg, and control mimic suffer from learning driving-irrelevant nuisances, perception errors, and not directly imbuing the representation with contextual knowledge, respectively.
\begin{table}[t!]
\centering
\vspace{-6pt}
\caption{Traffic light success rate (percentage of not running the {\em red} light). We compare with the best comparable baseline CILRS.}
\resizebox{0.65  \columnwidth}{!}{

\begin{tabular}{lccccc}
\toprule
              & \multicolumn{2}{c}{Train Town}                                    & \multicolumn{2}{c}{New Town}                                    &    \ \ \multirow{2}{*}{Mean}                      \\
Method          & \ \ \ Train weather & New weather & \ \ Train weather & New weather & \ \ \ \\ 
\midrule
CILRS~\cite{Codevilla_2019_ICCV} & \ \ \ $59\pm2$                            & $32\pm1$                             & \ \ $43\pm1$                             & $35\pm2$                            & \ \ \ $42\pm0.8$                       \\
SAM           & \ \ \ \textbf{97} $\pm$ \textbf{0}                            & \textbf{96} $\pm$ \textbf{1}                            & \ \ \textbf{81} $\pm$ \textbf{1}                            & \textbf{73} $\pm$ \textbf{1}                            & \ \ \ \textbf{87} $\pm$ \textbf{0.4}                       \\
\bottomrule
\end{tabular}
}
\vspace{-2mm}
\label{trafficlight}
\end{table}

\noindent \textbf{Traffic-light} \; Results are shown in Tab.~\ref{trafficlight}. As NoCrash does not directly penalize running red lights, \cite{Codevilla_2019_ICCV} proposes the Traffic-light benchmark to analyze traffic light violation behavior using the NoCrash empty setting (no dynamic agents). The metric is traffic-light success rate, i.e. the percentage of times the agent crosses a traffic-light on green. The traffic light success rate of {\em our model SAM} ($\textbf{87\%}$) is more than twice as high as {\em CILRS} ($\textbf{42\%}$). Our improved traffic light performance demonstrates the effectiveness of both our stop intentions and our method of leveraging them: the stop intentions inform the agent to stop for red lights while our method learns the context associated with them.

\subsection{A More Realistic Evaluation Protocol: Traffic-school}
\begin{table*}[t!]
\centering
\vspace{-8pt}
\caption{The newly proposed \textbf{\em Traffic-school} benchmark provides a more solid evaluation standard than both the CARLA, Supp. Mat., and NoCrash, Tab.~\ref{nocrash}, benchmarks, which are flawed due to not penalizing infractions such as red light or out-of-road violations. On our new benchmark, a route is considered successful only if the agent arrives at the destination within a given time  without a) crashing, b) traffic light violation, c) out-of-road infraction. Under this more realistic evaluation protocol, our results, in all conditions, surpass the best comparable baseline CILRS.}
\resizebox{1.0 \textwidth}{!}{

\begin{tabular}{lccclccclccclcccll}
\toprule
              & \multicolumn{3}{c}{\ Training} & \multicolumn{3}{c}{\ New weather} &  \multicolumn{3}{c}{\ New town} &  \multicolumn{3}{c}{\ New town/weather} &  \multirow{2}{*}{\ Mean}                         \\
Method        & \ \ Empty       & Regular      & Dense      & \ \ Empty       & Regular      & Dense      & \ \ Empty       & Regular      & Dense      & \ \ Empty       & Regular      & Dense      & \ \ \\
\midrule
CILRS~\cite{Codevilla_2019_ICCV} & \ \ $11\pm2$        & $12\pm1$         & $13\pm2$       & \ \ \ \ $2\pm2$         & \ \ $7\pm2$          & \ \ $7\pm1$        &  \ \ \ \ $2\pm1$         & \ \ \ $7\pm1$          & \ \ $4\pm1$        &  \ \ \ $0\pm0$         & \ \ $7\pm1$          & \ \ $2\pm2$        & \ \ \ \ $6\pm0.4$                   \\
SAM           & \ \ \textbf{90} $\pm$ \textbf{2}        & \textbf{79} $\pm$ \textbf{1}         & \textbf{43} $\pm$ \textbf{5}       & \ \ \textbf{83} $\pm$ \textbf{3}        & \textbf{73} $\pm$ \textbf{1}         & \textbf{39} $\pm$ \textbf{4}       &  \ \ \textbf{46} $\pm$ \textbf{2}        & \ \textbf{39} $\pm$ \textbf{3}         & \textbf{12} $\pm$ \textbf{2}       & \ \ \textbf{15} $\pm$ \textbf{3}        & \textbf{25} $\pm$ \textbf{2}         & \textbf{14} $\pm$ \textbf{0}       &  \ \ \textbf{47} $\pm$ \textbf{0.8}                   \\ 
\bottomrule
\end{tabular}
\vspace{-20pt}
}
\vspace{-2mm}
\label{noviolation}
\end{table*}

To resolve the flaws of previous benchmarks, we propose the {\em Traffic-school} benchmark, which shares the same routes and weathers as NoCrash with a more restrictive evaluation protocol. In previous benchmarks, multiple driving infractions such as red light violations (ignored by NoCrash) and out-of-lane violations are ignored when judging whether a route is successfully finished. In the {\em Traffic-school} benchmark, we do not ignore such infractions; a route is considered a success only if the agent reaches the destination while satisfying the following requirements: a) no overtime, b) no crashes, c) no traffic light violation, d) no running into the opposite lane or sidewalk. As shown in Tab.~\ref{noviolation}, under this more realistic evaluation protocol, our results (\textbf{47\%}) outperform the best comparable baseline CILRS (\textbf{6\%}), showing that our method learns the driving-related context and not the nuisances associated with side tasks. In particular, effectively leveraging the semantic segmentation and stop intentions boosts our performance for staying in the lane and stopping for red lights (Tab.~\ref{trafficlight}), infractions previously ignored but tested by {\em Traffic-school}.

\subsection{Ablation Studies about Squeeze-and-Mimic Networks}

To demonstrate the effectiveness of our method of using side task annotations, we conduct an ablation study on different methods of leveraging side tasks for driving (Tab.~\ref{advantageMD}). We compare against alternative methods for driving policy learning \cite{Muller2018DPT,Bansal2019ChaffeurNet,Xu2016EndtoEndLO,Li2018RethinkingSM,Hou2019FeatureMimick}, depicted in Fig.~\ref{fig:diffstrategies}, using the same side tasks and comparable backbones. We also compare against baselines that do not use side tasks. In the Supp. Mat., we visualize saliency maps to qualitatively show the advantages of our method.

\noindent \textbf{Two-stage-(F)} \; We apply an intuitive strategy (Fig.~\ref{fig:two-stage}) of utilizing two separately trained modules: a) perception networks, b) driving networks. The perception networks are trained for segmentation masks and stop intention values estimation. In the second step, the driving networks use the same architecture as the squeeze network and take estimated segmentation masks and stop intentions as input for low-level controls estimation. For two-stage, we directly take the learned weights from the squeeze network as the driving network. Note that the squeeze network is trained with ground-truth segmentation masks and stop intention values. Thus, for two-stage-F, we fine-tune the driving network on the estimated segmentation masks and stop intentions. Using either variant, we note that this strategy suffers from perception errors, which cannot be recovered from.

\noindent \textbf{Multi-task} \; We apply a similar multi-task training strategy (Fig.~\ref{fig:multi-task}) as \cite{Xu2016EndtoEndLO} and  MT~\cite{Li2018RethinkingSM} but with our side tasks. On the same latent feature vectors where we enforce mimicking losses in our SAM method, we now train decoders to estimate segmentation masks and stop intentions as side tasks. The motivation is that by simultaneously supervising these side tasks, the learned features contain driving-relevant context such as lane markings and other agents and are more invariant to environmental changes like buildings, weather, etc. However, the downside of this side task supervision is that it encourages the learned features to contain side task-relevant but driving-irrelevant nuisances.

\noindent \textbf{Feature mimic} Inspired by \cite{Hou2019FeatureMimick}, we construct a feature mimicking baseline (Fig.~\ref{fig:feature-mimick}) using our side tasks. Instead of mimicking the embeddings of a squeeze network trained to drive (SAM), we now mimic the embeddings of networks trained for semantic segmentation and stop intentions estimation. Similar to multi-task learning, feature mimicking also imbues the learned features with driving-relevant context but suffers from learning side task-relevant but driving-irrelevant nuisances.

\noindent \textbf{No mimicking} \;
We also compare against two baselines that do not use side tasks, SAM-NM and Res101-NM, to analyze the impact of effectively leveraging the side tasks via our method. For both models, we simply use $\ell_1$ losses for estimating low-level controls without enforcing the mimicking losses. SAM-NM uses the same two-ResNet architecture as SAM. 
As using two separate branches to process the image may be suboptimal in the no-mimicking case, we also compare to Res101-NM, a single-ResNet baseline that has a comparable number of network parameters. 

\begin{table}[t!]
\centering
\caption{Comparison of alternative methods that leverage the side task on NoCrash. We show navigation success rate in the new town. Training town results are included in Supp. Mat. Though two-stage-(F), multi-task, and feature mimic improve over the aforementioned non-distillation models by large gaps, they still perform worse than our SAM model, showing that among multiple alternatives of using the segmentation masks and stop intention values, our method performs best.}
\vspace{-6pt}
\resizebox{0.75 \textwidth}{!}{

\begin{tabular}{lccclcccc}
\toprule
            & \multicolumn{3}{c}{Training weather} &  & \multicolumn{3}{c}{New weather} &  \ \ \multirow{2}{*}{Mean} \\
Method      & \ \ \ Empty       & Regular      & Dense      & & Empty       & Regular      & Dense      &  \ \ \                   \\ 
\midrule
SAM-NM      & \ \ \ 65 $\pm$ 3        & 36 $\pm$ 1         & \ \ 9 $\pm$ 2        & & 42 $\pm$ 3        & 31 $\pm$ 2         & \ \ 7 $\pm$ 3        &   \ \ \ 31.7 $\pm$ 1.0             \\
Res101-NM   & \ \ \ 70 $\pm$ 3        & 44 $\pm$ 2         & 13 $\pm$ 4       & & 50 $\pm$ 2        & 33 $\pm$ 1         & \ \ 7 $\pm$ 3        &   \ \ \ 36.2 $\pm$ 1.1            \\
Two-stage   & \ \ \ \textbf{92} $\pm$ \textbf{1}        & 50 $\pm$ 3         & 12 $\pm$ 1       &  & 81 $\pm$ 2        & 41 $\pm$ 6         & \ \ 9 $\pm$ 3        &   \ \ \ 47.5 $\pm$ 1.3             \\
Two-stage-F & \ \ \ 90 $\pm$ 2        & 57 $\pm$ 4         & 13 $\pm$ 1       & & 79 $\pm$ 3        & 42 $\pm$ 4         & \ \ 8 $\pm$ 2        &   \ \ \ 48.2 $\pm$ 1.2            \\
Feature mimic & \ \ \ 90 $\pm$ 1 & 62 $\pm$ 2 & 18 $\pm$ 1 & & 79 $\pm$ 2 & 58 $\pm$ 2 & 15 $\pm$ 5 & \ \ \ 53.7 $\pm$ 1.0 \\
Multi-task  & \ \ \ 91 $\pm$ 0        & 62 $\pm$ 2         & 17 $\pm$ 2       & & \textbf{83} $\pm$ \textbf{1}        & 65 $\pm$ 6         & 16 $\pm$ 2       &   \ \ \ 55.7 $\pm$ 1.2             \\
SAM         & \ \ \ \textbf{92} $\pm$ \textbf{1}        & \textbf{74} $\pm$ \textbf{2}         & \textbf{29} $\pm$ \textbf{3}       & & \textbf{83} $\pm$ \textbf{1}        & \textbf{68} $\pm$ \textbf{7}         & \textbf{29} $\pm$ \textbf{2}       &   \ \ \ \textbf{62.5} $\pm$ \textbf{1.4}             \\ 
\bottomrule
\end{tabular}}
\label{advantageMD}
\vspace{-2mm}
\end{table}

\noindent From Tab.~\ref{advantageMD}, we see that our method outperforms the alternative approaches leveraging the side task. It outperforms multi-task and feature mimic, showing that it effectively learns a representation that contains only the driving-relevant information associated with the side task and not the nuisances. In addition, it outperforms two-stage-(F), avoiding the perception errors that plague two-stage methods. Finally, we note that all methods using the side tasks outperform the no-mimicking baselines, showing that our choice of side tasks (segmentation masks, stop intentions) is effective for improving generalization and providing driving-relevant context. 

\subsection{Ablation Studies about the Chosen Side Task Annotations}

We now conduct a careful ablation study to understand the impact of the chosen individual types of annotations. We analyze the influence of only utilizing one type of knowledge for mimicking: segmentation masks or stop intentions. In the Supp. Mat., we conduct ablation studies on the importance of each individual category of stop intention values, vehicle, pedestrian and traffic light, and show that the best performance is achieved when all three categories of intentions are used.

We use two different types of knowledge from the squeeze network for mimicking. Segmentation masks provide the mimic network with some simple concepts of object identities and therefore help the agent to learn basic driving skills like lane following and making turns. Meanwhile, stop intentions inform the agent of different hazardous traffic situations that require braking such as getting close to pedestrians, vehicles, or red lights. In Tab.~\ref{partial_mimic} we conduct ablation studies mimicking each type of information separately. Both types of knowledge separately bring performance gains, but the best results are achieved only when they are used jointly due to their complementary nature.

\begin{table}[t!]
\centering
\caption{Comparison of mimicking different types of knowledge. We show navigation success rate in the new town on NoCrash. Training town results are included in Supp. Mat. {\em Only stop intention} and {\em only segmentation mask} both improve upon the SAM-NM no-mimicking baseline, $31.7\%$, in Tab.~\ref{advantageMD}. The best results are achieved by mimicking both types of embedding knowledge jointly.}
\vspace{-6pt}
\resizebox{0.75 \textwidth}{!}{
\begin{tabular}{lccclccclc}
\toprule
                    & \multicolumn{3}{c}{Training weather} &  & \multicolumn{3}{c}{New weather} &  & \ \ \multirow{2}{*}{Mean} \\
Mimicking source & \ \ Empty   & Regular   & Dense   & \ \ \ & Empty   & Regular   & Dense   &  &                  \ \ \ \\ 
\midrule
Only stop intention & \ \ 86 $\pm$ 2    & 47 $\pm$ 3      & \ \ 8 $\pm$ 4     & \ \ \ & 73 $\pm$ 3    & 53 $\pm$ 6      & \ \ 9 $\pm$ 5     &  & \ \ \ 46.0 $\pm$ 1.7             \\
Only seg. mask      & \ \ \textbf{93} $\pm$ \textbf{1}    & 50 $\pm$ 4      & \ \ 8 $\pm$ 4     & \ \ \ & \textbf{85} $\pm$ \textbf{1}    & 52 $\pm$ 7      & \ \ 7 $\pm$ 3     &  & \ \ \ 49.2 $\pm$ 1.6             \\
Both                & \ \ 92 $\pm$ 1    & \textbf{74} $\pm$ \textbf{2}      & \textbf{29} $\pm$ \textbf{3}    & \ \ \ & 83 $\pm$ 1    & \textbf{68} $\pm$ \textbf{7}      & \textbf{29} $\pm$ \textbf{2}    &  & \ \ \ \textbf{62.5} $\pm$ \textbf{1.4}             \\ 
\bottomrule
\end{tabular}
}
\vspace{-2mm}
\label{partial_mimic}
\end{table}

\section{Discussion}

\label{sec:discussion}

We propose squeeze-and-mimic networks, a method that encourages the driving model to learn only driving-relevant context associated with a side task while discarding nuisances. We accomplish this by first squeezing the complementary side task annotations, semantic segmentation and stop intentions, using a squeeze network trained to drive. We then mimic this network's  representation to train the mimic network. Our method achieves state-of-the-art on various CARLA simulator benchmarks, including our newly proposed Traffic-school, which fixes previous benchmarks' flaws. In particular, {\em SAM} outperforms other approaches that also use side tasks.

However, our approach is not without limitations. Though it handles turning commands, it currently does not handle situations requiring negotiating with other agents such as lane changing and high-way merging, a potential topic for future work. Sensor fusion with LiDAR to further improve dense traffic performance could be another interesting direction.

\newpage

\acknowledgments{This work has been supported by grants ONR N00014-17-1-2072 and ARO W911NF-17-1-0304. This work has also been partially supported by NSF grants \#IIS-1633857, \#CCF-1837129, DARPA XAI grant
\#N66001-17-2-4032. We acknowledge the AWS Cloud Credits for Research program for providing computing resources.}

\bibliography{main}

\newpage

\appendix{

\etocdepthtag.toc{mtappendix}
\etocsettagdepth{mtchapter}{none}
\etocsettagdepth{mtappendix}{subsection}
\section*{Supplementary Material}

\tableofcontents

\label{sec:suppmat}

\vspace{-1pt}

\newpage

\section{Video Demo}

Please see the Supp. Video (\url{https://youtu.be/VRlHJTxf0Uc}), which illustrates the diversity of conditions (weather, number of agents) as well as the covariate shift between the training town and the new town, with representative successful runs of various maneuvers (stopping for  pedestrians, stopping at red light, turns with strong weather conditions, etc.). We also include a sample failure: colliding with vehicles while turning in dense traffic. We note that this failure mode appears rarely in the training dataset, and hence, our agent does not learn the proper behavior for this situation.

\section{Saliency Heatmaps}

To show qualitatively that our method is less sensitive to driving-irrelevant nuisances, we visualize saliency heatmaps (Fig.~\ref{fig:saliency}), generated by GradCAM~\cite{selvaraju2017grad}, for our SAM model and the multi-task model from our ablation study. We observe that our SAM model focuses its attention on driving-relevant entities like the road and sidewalk while the multi-task model focuses instead on driving-irrelevant nuisances such as the sky and buildings. These heatmaps, combined with the improved performance of our SAM method over multi-task learning (main paper Tab.~\ref{advantageMD}), demonstrate that our method learns a representation containing only driving-relevant information associated with the side task unlike multi-task learning, which overfits to these nuisances leading to worse performance.

\begin{figure}[t]
\ffigbox{
  \includegraphics[width=0.3\textwidth]{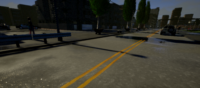}
  \includegraphics[width=0.3\textwidth]{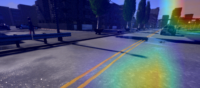}
  \includegraphics[width=0.3\textwidth]{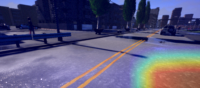}\\
  \vspace{+.5mm}
  \includegraphics[width=0.3\textwidth]{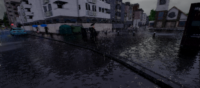}
  \includegraphics[width=0.3\textwidth]{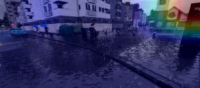}
  \includegraphics[width=0.3\textwidth]{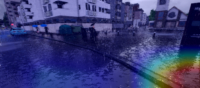}
}
   {\caption{Saliency heatmaps of the multi-task model (middle images) and our SAM model (right images) in the new town. The input images are on the left side. We see that our SAM model tends to focus on the road and sidewalk while the multi-task model tends to focus on driving-irrelevant nuisances such as the sky and buildings.}
   \label{fig:saliency}} 
\end{figure}

\section{Results on CARLA Benchmark}

The CARLA benchmark \cite{Dosovitskiy2017Carla} (Tab.~\ref{carla}) evaluates navigation with and without dynamic obstacles. The metric is navigation success rate, where a route is completed successfully if the agent reaches the destination within a time limit. This benchmark is relatively easy due to not penalizing crashes and therefore has been saturated. Nevertheless, {\em our model SAM} outperforms all comparable competing methods, achieving an average error of $\textbf{2\%}$ and a relative failure reduction of $\textbf{85\%}$ over the {\em second-best CILRS}, which achieves an average error of $\textbf{13\%}$.

\begin{table*}[t]
\centering
\caption{Results on CARLA benchmark ~\cite{Dosovitskiy2017Carla}. We show navigation success rate $(\%)$ in different test conditions. Dynamic / static indicate whether the test routes have moving objects (i.e. vehicles, pedestrians) or not. Bold indicates best among comparable methods. Italics indicate best among all methods, including methods such as LSD~\cite{ohn2020learning} and LBC~\cite{Chen2019lbc} whose results are provided solely for completeness. Note that LSD~\cite{ohn2020learning} reports results only on the new town. {\em Our model SAM} achieves an average error of $\textbf{2\%}$, surpassing the {\em second-best CILRS}, which has average error $\textbf{13\%}$, by $\textbf{85\%}$ in terms of relative failure reduction.}
\resizebox{1.0 \textwidth}{!}{
\begin{tabular}{lcclcclcclcclc}
\toprule
              & \multicolumn{2}{c}{Training} &  & \multicolumn{2}{c}{New weather} &  & \multicolumn{2}{c}{New town} &  & \multicolumn{2}{c}{New town/weather} &  & \ \ \ \multirow{2}{*}{Mean}    \\
Method        & \ \ \ Static             & Dynamic            &  & \ \ Static             & Dynamic            &  & \ \ Static             & Dynamic            &  &
\ Static             & Dynamic 
      &  &  \\ 
\midrule
\textbf{Comparable Baselines} \\
CIL~\cite{codevilla2018end}   & \ \ \ 86                 & 83                 &  & \ \ 84                 & 82                 &  & \ \ 40                 & 38                 &  & \ 44                 & 42                 &  & \ \ \ 62   \\
CAL~\cite{Sauer2018ConditionalAL}   & \ \ \ 92                 & 83                 &  & \ \ 90                 & 82                 &  & \ \ 70                 & 64                 &  & \ 68                 & 64                 &  & \ \ \ 77   \\
MT~\cite{Li2018RethinkingSM}    & \ \ \ 81                 & 81                 &  & \ \ 88                 & 80                 &  & \ \ 72                 & 53                 &  & \ 78                 & 62                 &  & \ \ \ 74   \\
CILRS~\cite{Codevilla_2019_ICCV} & \ \ \ 95                 & 92                 &  & \ \ 96                 & 96                 &  & \ \ 69                 & 66                 &  & \ 92                 & 90                 &  & \ \ \ 87   \\
SAM           & \ \ \ \textbf{100}                & \textbf{100}                &  & \ \ \textbf{100}                & \textit{\textbf{100}}                &  & \ \ \textbf{95}                 & \textbf{92}                 &  & \ \textbf{98}                 & \textbf{98}                 &  & \ \ \ \textbf{98}   \\ 
\textbf{Listed for Completeness} \\
LSD~\cite{ohn2020learning} & \ \ \ \ N/A                 & \ N/A                 &  & \ \ \ N/A                 & \ N/A                 &  & \ \ 99                 & 98                 &  & \ \ \textit{100}                 & 98                 &  & \ \ \ \ N/A   \\
LBC~\cite{Chen2019lbc} & \ \ \ \textit{100}               & \textit{100}                &  & \ \ \textit{100}                & 96 &  & \ \ \textit{100}                 & \textit{99}                &  & \ \textit{100}                 & \textit{100}                 &  & \ \ \ \textit{99}   \\
\bottomrule
\end{tabular}
}
\label{carla}
\vspace{-3mm}
\end{table*}

\section{Ablation Study about the Stop Intentions}

An autonomous driving agent might push its brake for various reasons, such as approaching other vehicles, pedestrians, or a red light. To analyze the impact of individual stop intentions on the learned driving model, we present Tab.~\ref{intention_types}. The results indicate that when all three types of intentions are used, the agent achieves the best performance as the set of all three stop intentions provides a complete causal link between braking and different hazardous traffic situations for the evaluation benchmark we use. Combined with the ablation study about the segmentation mask and stop intentions annotations (main paper section 4.4), we see that it is beneficial to jointly mimic both three-category stop intention embeddings and segmentation mask embeddings to achieve the best performance.

Furthermore, we observe that the only traffic light intentions model generally performs worse than the models using the other intentions. The relatively poor performance of this model could be due to the greater positive impact of the vehicle and pedestrian stop intentions during training compared to the traffic light intention. As the training data contains significant numbers of vehicles and pedestrians, the vehicle and pedestrian stop intentions may provide more benefit during training than the traffic light intention, leading to the performance gap between the traffic light intentions model and the other models.

\begin{table*}[t]
\centering
\vspace{-6pt}
\caption{Ablation study on three different categories of stop intention values, traffic light, vehicle and pedestrian, on NoCrash. We conduct ablation studies by mimicking the embeddings of squeeze networks trained using individual stop intentions. The best performance is achieved when all three-type stop intentions are used.}
\resizebox{1.0 \textwidth}{!}{

\begin{tabular}{lccclccclccclcccll}
\toprule
              & \multicolumn{3}{c}{\ Training} & \multicolumn{3}{c}{\ New weather} & \multicolumn{3}{c}{\ New town} & \multicolumn{3}{c}{\ New town/weather} &  \multirow{2}{*}{\ Mean}     \\
Stop intention          & Empty       & Regular      & Dense      & \ \ Empty       & Regular      & Dense      & \ \ Empty       & Regular      & Dense      &  \ \ Empty       & Regular      & \ Dense      & \ \  \\ 
\midrule
Traffic light & \ \ 95 $\pm$ 1    & 73 $\pm$ 1      & 16 $\pm$ 2  & \ \ 92 $\pm$ 0    & 63 $\pm$ 6      & \ \ 9 $\pm$ 1 &  \ \ 60 $\pm$ 3    & \ 37 $\pm$ 4      & 11 $\pm$ 3 & \ \ 46 $\pm$ 2    & 25 $\pm$ 4      & \ \ 9 $\pm$ 1 & \ \ $45\pm0.8$            \\
Vehicle & \textbf{100} $\pm$ \textbf{0}    & 89 $\pm$ 1      & 27 $\pm$ 3  & \ \ 94 $\pm$ 2    & 69 $\pm$ 1      & 25 $\pm$ 5     & \ \ 81 $\pm$ 2    & \ 50 $\pm$ 4      & 11 $\pm$ 2    &
  \ \ 73 $\pm$ 2    & 49 $\pm$ 7      & 13 $\pm$ 3   &  \ \ $57\pm0.9$            \\
Pedestrian & \textbf{100} $\pm$ \textbf{0}    & 93 $\pm$ 1      & 43 $\pm$ 2  & \ \ 98 $\pm$ 0    & 87 $\pm$ 5      & 41 $\pm$ 1    & \ \ 84 $\pm$ 2    & \ 61 $\pm$ 3      & 19 $\pm$ 1 &
 \ \ 71 $\pm$ 1    & 43 $\pm$ 1      & 13 $\pm$ 3  &  \ \ $63\pm0.6$            \\
All          & \textbf{100} $\pm$ \textbf{0}    & \textbf{94} $\pm$ \textbf{2}      & \textbf{54} $\pm$ \textbf{3}   & \ \textbf{100} $\pm$ \textbf{0}        & \textbf{89} $\pm$ \textbf{3}         & \textbf{47} $\pm$ \textbf{5}   &  \ \ \textbf{92} $\pm$ \textbf{1}    & \ \textbf{74} $\pm$ \textbf{2}      & \textbf{29} $\pm$ \textbf{3} & \ \ \textbf{83} $\pm$ \textbf{1}    & \textbf{68} $\pm$ \textbf{7}      & \textbf{29} $\pm$ \textbf{2}  &  \ \ $\textbf{72} \pm \textbf{0.9}$                   \\ 
\bottomrule
\end{tabular}
\vspace{-20 pt}
}
\label{intention_types}
\vspace{-6pt}
\end{table*}

\section{Additional Results on NoCrash Training Town}

\begin{table}[t]
\centering
\caption{Comparison of alternative methods that leverage the side task on NoCrash~\cite{Codevilla_2019_ICCV} in the training town. We show navigation success rate $(\%)$ in different weather conditions.} 
\vspace{-6pt}
\resizebox{0.75 \textwidth}{!}{
\begin{tabular}{lccclcccc}
\toprule
            & \multicolumn{3}{c}{Training weather} & & \multicolumn{3}{c}{New weather} &  \ \ \multirow{2}{*}{Mean} \\
Method \ \ \      & Empty       & Regular      & Dense      &  & Empty       & Regular      & Dense      & \ \ \                    \\ 
\midrule
SAM-NM \ \ \      & \ \ 98 $\pm$ 0        & 81 $\pm$ 2      & 19 $\pm$ 3        &  & 96 $\pm$ 0        & 72 $\pm$ 4         & 18 $\pm$ 5        &   \ \ \ 64.0 $\pm$  1.2         \\
Res101-NM \ \ \  & \ \ 99 $\pm$ 1        & 85 $\pm$ 2         & 22 $\pm$ 1       &  & 89 $\pm$ 1        & 72 $\pm$ 3         & 27 $\pm$ 1        &   \ \ \ 65.7 $\pm$ 0.7             \\
Two-stage \ \ \  & \textbf{100} $\pm$ \textbf{0}        & 80 $\pm$ 4         & 29 $\pm$ 4       &  & 83 $\pm$ 1        & 63 $\pm$ 1         & 15 $\pm$ 4        &   \ \ \ 61.7 $\pm$ 1.2             \\
Two-stage-F \ \ \ & \textbf{100} $\pm$ \textbf{0}        & 83 $\pm$ 1         & 29 $\pm$ 4       &  & 87 $\pm$ 1        & 67 $\pm$ 2         & 22 $\pm$ 5        &   \ \ \ 64.7 $\pm$ 1.1             \\
Feature mimic \ \ \ & \textbf{100} $\pm$ \textbf{0} & 87 $\pm$ 1 & 34 $\pm$ 3 & & \textbf{100} $\pm$ \textbf{0} & 83 $\pm$ 5 & 34 $\pm$ 6 & \ \ \ 73.0 $\pm$ 1.4 \\
Multi-task \ \ \ & \textbf{100} $\pm$ \textbf{0}        & \textbf{94} $\pm$ \textbf{3}         & 41 $\pm$ 2       &  & 96 $\pm$ 0        & 87 $\pm$ 2         & 37 $\pm$ 5       &   \ \ \ 75.8 $\pm$ 1.1             \\
SAM  \ \ \        & \textbf{100} $\pm$ \textbf{0}        & \textbf{94} $\pm$ \textbf{2}         & \textbf{54} $\pm$ \textbf{3}       &  & \textbf{100} $\pm$ \textbf{0}        & \textbf{89} $\pm$ \textbf{3}         & \textbf{47} $\pm$ \textbf{5}       &   \ \ \ \textbf{80.7} $\pm$ \textbf{1.1}            \\ \bottomrule
\end{tabular}}
\label{advantageMDtownA}
\vspace{-6pt}
\end{table}

\begin{table}[t]
\centering
\caption{Comparison of mimicking different types of knowledge on NoCrash in the training town. We show navigation success rate $(\%)$ in different weathers.}
\vspace{-6pt}
\resizebox{0.75 \columnwidth}{!}{

\begin{tabular}{lccclccclc}
\toprule
                    & \multicolumn{3}{c}{Training weather} &  & \multicolumn{3}{c}{New weather} &  & \ \ \multirow{2}{*}{Mean} \\
Mimicking source \ \ & Empty   & Regular   & Dense   & \ \ \  & Empty   & Regular   & Dense   &  &                \ \ \ \\ 
\midrule
Only stop intention \ \ & \textbf{100} $\pm$ \textbf{0}    & 86 $\pm$ 2      & 33 $\pm$ 4     & \ \ \  & \ \ 97 $\pm$ 1    & 79 $\pm$ 1      & 25 $\pm$ 2     &  & \ \ \  70.0 $\pm$ 0.8             \\
Only seg. mask   \ \   & \textbf{100} $\pm$ \textbf{0}    & 83 $\pm$ 3      & 31 $\pm$ 3     & \ \ \  & \ \ 98 $\pm$ 2    & 79 $\pm$ 5      & 23 $\pm$ 6     &  & \ \ \  69.0 $\pm$ 1.5             \\
Both          \ \       & \textbf{100} $\pm$ \textbf{0}    & \textbf{94} $\pm$ \textbf{2}      & \textbf{54} $\pm$ \textbf{3}    & \ \ \  & \textbf{100} $\pm$ \textbf{0}        & \textbf{89} $\pm$ \textbf{3}         & \textbf{47} $\pm$ \textbf{5}     &  & \ \ \ \textbf{80.7} $\pm$ \textbf{1.1}             \\ 
\bottomrule
\end{tabular}

}
\label{partial_mimictownA}
\vspace{-6pt}
\end{table}

\begin{table*}[t!]
\centering
\vspace{-6pt}
\caption{Comparison of alternative methods that leverage the side task on Traffic-school. We show navigation success rate $(\%)$ in different test conditions.}
\resizebox{1.0 \textwidth}{!}{

\begin{tabular}{lccclccclccclcccll}
\toprule
                & \multicolumn{3}{c}{\ Training} & \multicolumn{3}{c}{\ New weather} & \multicolumn{3}{c}{\ New town} & \multicolumn{3}{c}{\ New town/weather} &  \multirow{2}{*}{\ \ Mean}     \\
Method          & Empty       & Regular      & Dense      & \ \ Empty       & Regular      & Dense      & \ \ Empty       & Regular      & Dense      &  \ \ Empty       & Regular      & Dense      & \ \  \\ 
\midrule
SAM-NM & $32\pm1$        & $26\pm2$         & $10\pm3$       & \ \ $35\pm3$         & $21\pm4$          & \ \ $7\pm4$        & \ \ \ \ $9\pm0$         & \ \ \ $7\pm1$          & \  $3\pm1$        & \ \ \ $4\pm0$         & \ \ $9\pm3$          & \ $3\pm2$        &  \ \ \ $14\pm0.7$                   \\
Res101-NM & $35\pm2$        & $27\pm2$         & $15\pm1$       & \ \ $26\pm2$         & $16\pm2$          & $16\pm2$        & \ \  $14\pm2$         & \ $13\pm3$          & \  $6\pm2$        &  \ \ \textbf{17} $\pm$ \textbf{1}         & $14\pm2$          &  \ $3\pm1$        &  \ \ \ $17\pm0.6$                   \\
Two-stage & $16\pm0$        & $12\pm2$         & $14\pm1$       & \ \ \ \ $9\pm1$      & \ \ $7\pm1$       & \ \ $4\pm2$         & \ \ \ \ $2\pm1$          & \ \ \ $9\pm4$           & \ $2\pm1$ & \ \ \ $5\pm2$          & \ \ $2\pm0$       & \ $2\pm2$   & \ \ \ \ \ $7\pm0.5$            \\
Two-stage-F & $19\pm2$        & $14\pm1$         & $18\pm2$       &  \ \ \ \ $7\pm1$      & \ \ $8\pm3$       & \ \ $9\pm2$         & \ \ \ \ $3\pm1$          & \ $12\pm3$           & \  $4\pm1$ & \ \ \ $5\pm2$          & \ \ $1\pm1$       & \ $1\pm2$   &  \ \ \ \ \ $8\pm0.5$            \\
Feature mimic & $22\pm1$ & $20\pm1$ & $19\pm4$ & \ \ $17\pm1$ & $17\pm3$ & $16\pm3$ & \ \ $11\pm1$ & \ $13\pm1$ & \ $8\pm2$ & \ \ $10\pm0$ & $15\pm1$ & \ $9\pm3$  & \ \ \ $15\pm0.6$ \\  
Multi-task & $83\pm1$        & $72\pm3$         & $32\pm1$       &  \ \ $62\pm0$      & $57\pm3$       & $28\pm2$         & \ \ $17\pm1$          & \ $16\pm1$           & \ $6\pm0$ & \ \ \ $9\pm1$          & $10\pm2$       & \ $5\pm1$   &  \ \ \ $33\pm0.5$            \\
SAM           & \textbf{90} $\pm$ \textbf{2}        & \textbf{79} $\pm$ \textbf{1}         & \textbf{43} $\pm$ \textbf{5}       & \ \  \textbf{83} $\pm$ \textbf{3}        & \textbf{73} $\pm$ \textbf{1}         & \textbf{39} $\pm$ \textbf{4}       & \ \  \textbf{46} $\pm$ \textbf{2}        & \ \textbf{39} $\pm$ \textbf{3}         & \textbf{12} $\pm$ \textbf{2}       &  \ \ 15 $\pm$ 3        & \textbf{25} $\pm$ \textbf{2}         & \textbf{14} $\pm$ \textbf{0}       & \ \ \ $\textbf{47} \pm \textbf{0.8}$                   \\ 
\bottomrule
\end{tabular}
\vspace{-20 pt}
}
\label{advantageMDnoviolation}
\vspace{-6pt}
\end{table*}

For the tables where only new town results were provided on NoCrash, here we demonstrate training town results. Tab.~\ref{advantageMDtownA} and Tab.~\ref{partial_mimictownA}  provide the training town results corresponding to Tab.~\ref{advantageMD} and \ref{partial_mimic} in the main paper, respectively. We observe from Tab.~\ref{advantageMDtownA}, that our model outperforms the alternative methods of leveraging the side task in the training town, consistent with the new town results. We also see from Tab.~\ref{partial_mimictownA} that mimicking both types of knowledge from the segmentation mask and stop intention values jointly provides the best performance in the training town, consistent with the new town results.

We now discuss an interesting observation from the ablation study for mimicking different types of knowledge (main paper Tab.~\ref{partial_mimic}). We observe that in the Empty setting, the model mimicking only segmentation mask embedding outperforms the model mimicking only stop intentions embedding. This trend can be explained by considering the different types of knowledge segmentation mask and stop intentions provide. Segmentation mask provides class knowledge for tasks such as lane following while stop intentions provide causal braking indicators for tasks such as collision avoidance. In the Empty setting, the agent does not need to stop for vehicles or pedestrians but still needs to follow the lane and make turns, so the braking knowledge provided by the stop intentions is not that important in this setting compared to the class knowledge provided by the segmentation mask. Hence, the only segmentation mask embedding model performs better in Empty compared to the only stop intentions embedding model.

\section{Additional Results on Traffic-school}

\begin{table*}[t]
\centering
\vspace{-6pt}
\caption{Comparison of mimicking different knowledge types on Traffic-school. We show navigation success rate $(\%)$ in different test conditions.}
\resizebox{1.0 \textwidth}{!}{

\begin{tabular}{lccclccclccclcccll}
\toprule
                & \multicolumn{3}{c}{\ Training} & \multicolumn{3}{c}{\ New weather} & \multicolumn{3}{c}{\ New town} & \multicolumn{3}{c}{\ New town/weather} &  \multirow{2}{*}{\ \ Mean}     \\
Mimicking source          & Empty       & Regular      & Dense      & \ \ Empty       & Regular      & Dense      & \ \ Empty       & Regular      & Dense      &  \ \ \ Empty       & Regular      & \ Dense      & \ \  \\ 
\midrule
Only stop intention & $28\pm3$        & $19\pm2$         & $22\pm5$       & \ \ $21\pm1$      & $17\pm2$       & $10\pm3$         &  \ \ \ \ $7\pm1$          & \ \ $8\pm2$           & \ \ $2\pm1$ & \ \ \ \ $6\pm2$          & $11\pm3$       & \ \ $4\pm0$   & \ \ \ $13\pm0.7$            \\
Only seg. mask & $36\pm3$        & $28\pm3$         & $16\pm3$       & \ \ $23\pm3$      & $26\pm0$       & $11\pm3$         & \ \ \ \ $6\pm1$          & \ $10\pm1$           & \ \ $2\pm1$ & \ \ \ \ $7\pm3$          & $11\pm5$       & \ \ $5\pm1$   &  \ \ \ $15\pm0.8$            \\
Both           & \textbf{90} $\pm$ \textbf{2}        & \textbf{79} $\pm$ \textbf{1}         & \textbf{43} $\pm$ \textbf{5}       & \ \  \textbf{83} $\pm$ \textbf{3}        & \textbf{73} $\pm$ \textbf{1}         & \textbf{39} $\pm$ \textbf{4}       & \ \  \textbf{46} $\pm$ \textbf{2}        & \ \textbf{39} $\pm$ \textbf{3}         & \textbf{12} $\pm$ \textbf{2}       & \ \  \textbf{15} $\pm$ \textbf{3}        & \textbf{25} $\pm$ \textbf{2}         & \textbf{14} $\pm$ \textbf{0}       & \ \ \ $\textbf{47} \pm \textbf{0.8}$                   \\ 
\bottomrule
\end{tabular}
\vspace{-20 pt}
}
\label{partial_mimicnoviolation}
\vspace{-6pt}
\end{table*}

\begin{table*}[t!]
\centering
\vspace{-6pt}
\caption{Ablation study on three different categories of stop intention values on Traffic-school. We show navigation success rate $(\%)$ in different test conditions.}
\resizebox{1.0 \textwidth}{!}{

\begin{tabular}{lccclccclccclcccll}
\toprule
                & \multicolumn{3}{c}{\ Training} & \multicolumn{3}{c}{\ New weather} & \multicolumn{3}{c}{\ New town} & \multicolumn{3}{c}{\ New town/weather} &  \multirow{2}{*}{\ \ Mean}     \\
Stop intention          & Empty       & Regular      & Dense      & \ \ Empty       & Regular      & Dense      & \ \ Empty       & Regular      & \ Dense      &  \ \ Empty       & \ Regular      & \ Dense      & \ \  \\ 
\midrule
Traffic light & $34\pm1$        & $21\pm3$         & $11\pm2$       &  \ \ $26\pm0$      & $10\pm3$       & \ \ $3\pm3$         &  \ \ \ \ $9\pm0$          & \ $11\pm3$           & \ \ $5\pm1$ & \ \ \ \ $6\pm2$          & \ \ $9\pm1$       & \ \ $5\pm1$   & \ \ \ $13\pm0.6$            \\
Vehicle & $36\pm1$        & $30\pm1$         & $11\pm2$       & \ \ $20\pm2$      & $17\pm3$       & \ \ $5\pm1$         & \ \ \ \ $5\pm1$          & \ \ \ $7\pm2$           & \ \ $1\pm2$ & \ \ \ \ $5\pm1$          & \ \ $6\pm2$       & \ \ $4\pm2$   & \ \ \ $12\pm0.5$            \\
Pedestrian & $55\pm1$        & $51\pm2$         & $29\pm1$       & \ \ $54\pm0$      & $48\pm5$       & $23\pm5$         & \ \ $10\pm2$          & \ $10\pm2$           & \ \ $3\pm2$ & \ \ \ \ $7\pm3$          & \ \ $5\pm1$       & \ \ $2\pm0$   & \ \ \ $25\pm0.7$            \\
All          & \textbf{90} $\pm$ \textbf{2}        & \textbf{79} $\pm$ \textbf{1}         & \textbf{43} $\pm$ \textbf{5}       & \ \  \textbf{83} $\pm$ \textbf{3}        & \textbf{73} $\pm$ \textbf{1}         & \textbf{39} $\pm$ \textbf{4}       & \ \  \textbf{46} $\pm$ \textbf{2}        & \ \textbf{39} $\pm$ \textbf{3}         & \textbf{12} $\pm$ \textbf{2}       & \ \  \textbf{15} $\pm$ \textbf{3}        & \textbf{25} $\pm$ \textbf{2}         & \textbf{14} $\pm$ \textbf{0}       & \ \ \ $\textbf{47} \pm \textbf{0.8}$                   \\ 
\bottomrule
\end{tabular}
\vspace{-20 pt}
}
\label{intention_typesnoviolation}
\vspace{-6pt}
\end{table*}

We also provide results for the ablation studies on our newly proposed Traffic-school benchmark. Tab.~\ref{advantageMDnoviolation},~\ref{partial_mimicnoviolation},~\ref{intention_typesnoviolation} provide the Traffic-school results (both training and new towns) corresponding to main paper Tab.~\ref{advantageMD}, \ref{partial_mimic}, and Supp Mat Tab.~\ref{intention_types}, respectively. We see that on the Traffic-school benchmark, similar to the results on NoCrash, our model outperforms all alternative methods of leveraging the side task as it avoids perception errors (\textit{e.g.} errors in segmentation mask and stop intentions estimation) that plague two-stage methods and learns only driving-relevant context squeezed by the squeeze network from the side task annotations unlike multi-task learning and feature mimicking. We also observe that both our method and multi-task learning outperform the no-mimicking methods on Traffic-school. This demonstrates the effectiveness of our choice of side tasks: semantic segmentation provides object class identity, aiding with basic tasks such as lane following, while three-category stop intentions provide causal information relating braking to various stopping causes. Furthermore, similar to the results on NoCrash, we observe that jointly mimicking both the segmentation mask and \emph{three-category} stop intentions embeddings leads to the best performance on Traffic-school, demonstrating the complementary nature of the segmentation mask and three-category stop intentions side tasks. 

\section{Comparison with Squeeze Network}

\begin{table*}[t!]
\centering
\vspace{-6pt}
\caption{Comparison of SAM and squeeze network on the NoCrash benchmark. We show navigation success rate $(\%)$ in different test conditions.}
\resizebox{1 \textwidth}{!}{

\begin{tabular}{lccclccclccclccclc}
\toprule
                & \multicolumn{3}{c}{\ Training} & \multicolumn{3}{c}{\ New weather} & \multicolumn{3}{c}{\ New town} & \multicolumn{3}{c}{\ New town/weather} &  \multirow{2}{*}{\ Mean}     \\
Method          & \ \ Empty       & Regular      & Dense      & \ \ \ Empty       & Regular      & Dense      & \ Empty       & Regular      & Dense      &  \ \ Empty       & Regular      & Dense      & \\ 
\midrule
SAM             & \ \ \textbf{100} $\pm$ \textbf{0}         & \textbf{94} $\pm$ \textbf{2}          & 54 $\pm$ 3        & \ \  \textbf{100} $\pm$ \textbf{0}         & 89 $\pm$ 3          & 47 $\pm$ 5        & \ \ 92 $\pm$ 1         & \ 74 $\pm$ 2          & 29 $\pm$ 3        &  \ \ 83 $\pm$ 1         & 68 $\pm$ 7          & 29 $\pm$ 2        &  \ 72 $\pm$ 0.9   \\ 
Squeeze            & \ \ \textbf{100} $\pm$ \textbf{0} &	93 $\pm$ 2 & \textbf{63} $\pm$ \textbf{7}	& \ \ \textbf{100} $\pm$ \textbf{0}	& \textbf{93} $\pm$ \textbf{2} & \textbf{59} $\pm$ \textbf{4} & \ \ 	\textbf{97} $\pm$ \textbf{1} & \ \textbf{76} $\pm$ {3} & \textbf{40} $\pm$ \textbf{4} & \ \  \textbf{99} $\pm$ \textbf{2} & \textbf{81} $\pm$ \textbf{3} &	\textbf{39} $\pm$ \textbf{1} & \ \textbf{78} $\pm$ \textbf{0.9}  \\
\bottomrule
\end{tabular}

}
\label{expertnocrash}
\vspace{-6 pt}
\end{table*}

\begin{table*}[t!]
\centering
\vspace{-6pt}
\caption{Comparison of SAM and squeeze network on the Traffic-school benchmark. We show navigation success rate $(\%)$ in different test conditions.}
\resizebox{1.0 \textwidth}{!}{

\begin{tabular}{lccclccclccclcccll}
\toprule
& \multicolumn{3}{c}{\ Training} & \multicolumn{3}{c}{\ New weather} & \multicolumn{3}{c}{\ New town} & \multicolumn{3}{c}{\ New town/weather} &  \multirow{2}{*}{\ Mean}     \\
Method          & \ \ Empty       & Regular      & Dense      & \ \ Empty       & Regular      & Dense      & \ \ Empty       & Regular      & Dense      &  \ \ Empty       & Regular      & Dense      & \\ 
\midrule                
SAM           & \ \ \textbf{90} $\pm$ \textbf{2}        & \textbf{79} $\pm$ \textbf{1}         & 43 $\pm$ 5       &  \ \ \textbf{83} $\pm$ \textbf{3}        & \textbf{73} $\pm$ \textbf{1}         & 39 $\pm$ 4       & \ \  \textbf{46} $\pm$ \textbf{2}        & \ 39 $\pm$ 3         & 12 $\pm$ 2       &  \ \ 15 $\pm$ 3        & 25 $\pm$ 2         & 14 $\pm$ 0       &  \ 47 $\pm$ 0.8                   \\ 
Squeeze          & \ \ 76 $\pm$ 1        & 61 $\pm$ 1         & \textbf{45} $\pm$ \textbf{4}      & \ \ 75 $\pm$ 2        & 61 $\pm$ 4         & \textbf{45} $\pm$ \textbf{10}       & \ \  39 $\pm$ 1        & \ \textbf{40} $\pm$ \textbf{2}         & \textbf{23} $\pm$ \textbf{4}       & \ \ \textbf{39} $\pm$ \textbf{3}        & \textbf{43} $\pm$ \textbf{1}         & \textbf{23} $\pm$ \textbf{1}       & \ \textbf{48} $\pm$ \textbf{1.1} \\
\bottomrule
\end{tabular}

}
\label{expertnoviolation}
\vspace{-6 pt}
\end{table*}

We compare to the squeeze network in Tab.~\ref{expertnocrash} and Tab.~\ref{expertnoviolation} to evaluate how well our SAM model (mimic network) mimics the squeeze network's embedding for the purpose of driving. We note that this comparison is not exactly fair as the squeeze network has access to ground-truth semantic segmentation and stop intention values at test-time unlike our SAM model, which has access to only images. Since we assume that the squeeze network's ground-truth inputs are generally unavailable at test time, for this comparison, we provide these inputs to the squeeze network. 

As expected, the squeeze network, which has access to the ground-truth semantic segmentation and stop intention values at test time, generally outperforms the SAM model. However, the SAM model does not lag too far behind compared to the squeeze network (6\% worse in success rate on NoCrash and comparable performance on Traffic-school), showing that the SAM model effectively mimics the squeeze network's embedding. The SAM model will occasionally outperform the squeeze network due to overfitting to photometric nuisances in the training town. As a consequence, our SAM model does not generalize as well to the new town compared to the squeeze network, which has access to ground truth side task annotations.

\section{Ablation Study on Amount of Side Task Annotations}
To examine the performance of our SAM method with varying amounts of side task annotations, we also train our SAM method using 0 hours (SAM-NM), 5 hours, and 7 hours of annotations (semantic segmentation and stop intentions) in addition to the default 10 hours. The results on NoCrash and Traffic-school are in Tab.~\ref{annotationamount} and ~\ref{annotationamountnoviolation}, respectively. We observe that leveraging any amount (5 hours, 7 hours, or 10 hours) of side task annotations leads to significant performance gains over using no side task annotations, showing the effectiveness of our method even if side task annotations are not present for all frames, a situation that may occur in practice. Furthermore, we note that using 7 hours of side task annotations leads to comparable performance as using the full 10 hours of annotations, showing that our SAM method is robust to missing side task annotations in the dataset. Finally, we note that our SAM method (SAM 5 hours), despite using only half as many side task annotations, performs comparably to multi-task learning on NoCrash (67\%, Tab.~\ref{annotationamount} vs. 66\%, main paper Tab.~\ref{advantageMD} and Supp Mat Tab.~\ref{advantageMDtownA}) and outperforms multi-task learning on Traffic-school (43\%, Tab.~\ref{annotationamountnoviolation} vs. 33\%, Tab.~\ref{advantageMDnoviolation}). As multi-task learning is the second-best method of leveraging the side tasks (main paper Tab.~\ref{advantageMD}), this demonstrates that our method, among all the alternatives, most effectively leverages the side task as it learns a representation that contains driving-relevant context from the side task without the associated driving-irrelevant nuisances.

\begin{table*}[t!]
\centering
\vspace{-6pt}
\caption{Comparison of SAM with differing amounts of side task annotations on NoCrash. We show navigation success rate $(\%)$ in different test conditions.}
\resizebox{1 \textwidth}{!}{

\begin{tabular}{lccclccclccclccclc}
\toprule
& \multicolumn{3}{c}{\ Training} & \multicolumn{3}{c}{\ New weather} & \multicolumn{3}{c}{\ New town} & \multicolumn{3}{c}{\ New town/weather} &  \multirow{2}{*}{\ Mean}     \\
Method          & Empty       & Regular      & Dense      & \ \ Empty       & Regular      & Dense      & \ \ Empty       & Regular      & Dense      &  \ \ Empty       & Regular      & Dense      & \ \  \\ 
\midrule
SAM-NM (0 hours)          & \ 98 $\pm$ 0        & 81 $\pm$ 2      & 19 $\pm$ 3        & \ \  96 $\pm$ 0        & \ 72 $\pm$ 4         & 18 $\pm$ 5      &  \ \ 65 $\pm$ 3        & \ 36 $\pm$ 1         & \ \ 9 $\pm$ 2        & \ \ 42 $\pm$ 3        & 31 $\pm$ 2         & \ \ 7 $\pm$ 3        & \ \ \ 48 $\pm$ 0.8      \\
SAM 5 hours            & \textbf{100} $\pm$ \textbf{0} & \textbf{94} $\pm$ \textbf{1} & 53 $\pm$ 3 & \ \textbf{100} $\pm$ \textbf{0} & \ 91 $\pm$ 1 & 43 $\pm$ 7 & \ \ 85 $\pm$ 2 & \ 61 $\pm$ 3 & 24 $\pm$ 6 & \ \ 76 $\pm$ 2 & 56 $\pm$ 9 & 21 $\pm$ 2 & \ \ \ 67 $\pm$ 1.2 \\
SAM 7 hours            & \textbf{100} $\pm$ \textbf{0} &	93 $\pm$ 1 & 53 $\pm$ 5	& \ \ 99 $\pm$ 1	& \ \textbf{92} $\pm$ \textbf{4} & 43 $\pm$ 5 & \ \ \textbf{94} $\pm$ \textbf{0} & \ 72 $\pm$ 2 & \textbf{31} $\pm$ \textbf{4} & \ \ \textbf{83} $\pm$ \textbf{1} & \textbf{69} $\pm$ \textbf{1} & 27 $\pm$ 2 & \ \ \ 71 $\pm$ 0.8 \\
SAM 10 hours        & \textbf{100} $\pm$ \textbf{0}         & \textbf{94} $\pm$ \textbf{2}          & \textbf{54} $\pm$ \textbf{3}        &  \ \textbf{100} $\pm$ \textbf{0}         & \ 89 $\pm$ 3          & \textbf{47} $\pm$ \textbf{5}        &  \ \ 92 $\pm$ 1         & \ \textbf{74} $\pm$ \textbf{2}          & 29 $\pm$ 3        &  \ \ \textbf{83} $\pm$ \textbf{1}         & 68 $\pm$ 7          & \textbf{29} $\pm$ \textbf{2}        & \ \ \ \textbf{72} $\pm$ \textbf{0.9}   \\ 
\bottomrule
\end{tabular}

}
\vspace{-6pt}
\label{annotationamount}
\end{table*}

\begin{table*}[t!]
\centering
\vspace{-6pt}
\caption{Comparison of SAM with differing amounts of side task annotations on Traffic-school. We show navigation success rate $(\%)$ in different test conditions.}
\resizebox{1 \textwidth}{!}{

\begin{tabular}{lccclccclccclccclc}
\toprule
& \multicolumn{3}{c}{\ Training} & \multicolumn{3}{c}{\ New weather} & \multicolumn{3}{c}{\ New town} & \multicolumn{3}{c}{\ New town/weather} &  \multirow{2}{*}{\ Mean}     \\
Method          & Empty       & Regular      & Dense      & \ \ Empty       & Regular      & Dense      & \ \ Empty       & Regular      & Dense      &  \ \ Empty       & Regular      & Dense      & \ \  \\ 
\midrule
SAM-NM (0 hours)          & $32\pm1$        & $26\pm2$         & $10\pm3$       & \ \ $35\pm3$         & $21\pm4$          & \ \ $7\pm4$        & \ \ \ \ $9\pm0$         & \ \ $7\pm1$          & \  $3\pm1$        & \ \ \ $4\pm0$         & \ \ $9\pm3$          & \ $3\pm2$        &  \ \ \ $14\pm0.7$  \\
SAM 5 hours            & 84 $\pm$ 3 & 78 $\pm$ 1 & \textbf{44} $\pm$ \textbf{1} & \ \ 82 $\pm$ 3 & \textbf{74} $\pm$ \textbf{9} & 33 $\pm$ 6 & \ \ 40 $\pm$ 1 & \ 33 $\pm$ 1 & 12 $\pm$ 3 & \ \ 13 $\pm$ 1 & 13 $\pm$ 4 & \ 7 $\pm$ 1 & \ \ \ 43 $\pm$ 1.1 \\
SAM 7 hours            & \textbf{91} $\pm$ \textbf{1} & \textbf{82} $\pm$ \textbf{1} & 42 $\pm$ 4	& \ \ 81 $\pm$ 1	& \textbf{74} $\pm$ \textbf{5} & 36 $\pm$ 7 & \ \ 42 $\pm$ 3 & \ 36 $\pm$ 1 & \textbf{16} $\pm$ \textbf{2} & \ \ \textbf{23} $\pm$ \textbf{1} & \textbf{25} $\pm$ \textbf{2} & 10 $\pm$ 2 & \ \ \ \textbf{47} $\pm$ \textbf{0.9} \\
SAM 10 hours        & 90 $\pm$ 2        & 79 $\pm$ 1         & 43 $\pm$ 5       &  \ \ \textbf{83} $\pm$ \textbf{3}        & 73 $\pm$ 1         & \textbf{39} $\pm$ \textbf{4}       & \ \  \textbf{46} $\pm$ \textbf{2}        & \ \textbf{39} $\pm$ \textbf{3}         & 12 $\pm$ 2       &  \ \ 15 $\pm$ 3        & \textbf{25} $\pm$ \textbf{2}         & \textbf{14} $\pm$ \textbf{0}       &  \ \ \ \textbf{47} $\pm$ \textbf{0.8}      \\ 
\bottomrule
\end{tabular}

}
\vspace{-6pt}
\label{annotationamountnoviolation}
\end{table*}

\section{Ablation Study on Mimic Loss Hyperparameters}

We conduct an ablation study (Tab.~\ref{lossweightsablation} and \ref{lossweightsablationnoviolation}) on the mimic loss weights ($\lambda_1^{'}, \lambda_2^{'}$ in main paper equation~\ref{latent_loss}) to examine the sensitivity of our SAM method with regards to these weights; note that our method uses $\lambda_1^{'} = 0.03, \lambda_2^{'} = 0.03$, the setting that performs best. We see that our method is fairly robust to various hyper-parameter settings as long as both mimic loss terms are used ($\lambda_1^{'}, \lambda_2^{'} > 0$). Furthermore, we observe that performance decreases significantly with $\lambda_1^{'} = 0$ or $\lambda_2^{'} = 0$, showing that it is important to mimic both segmentation and stop intentions knowledge from the squeeze network.

\begin{table*}[t]
\centering
\vspace{-6pt}
\caption{Ablation study on mimic loss weights ($\lambda_1^{'}, \lambda_2^{'}$) on NoCrash. We show navigation success rate $(\%)$ in different test conditions. We bold ($\lambda_1^{'}, \lambda_2^{'}$) = (0.03, 0.03) as this set of weights performs the best and is the one we use.}
\resizebox{1 \textwidth}{!}{
\begin{tabular}{lccclccclccclccclc}
\toprule
& \multicolumn{3}{c}{\ Training} & \multicolumn{3}{c}{\ New weather} & \multicolumn{3}{c}{\ New town} & \multicolumn{3}{c}{\ New town/weather} &  \multirow{2}{*}{\ Mean}     \\
($\lambda_1^{'}, \lambda_2^{'}$)          & Empty       & Regular      & Dense      & \ \ Empty       & Regular      & Dense      & \ \ Empty       & Regular      & Dense      &  \ \ Empty       & Regular      & Dense      & \ \  \\ 
\midrule
(0, 0.03)          & \textbf{100} $\pm$ \textbf{0}    & 86 $\pm$ 2      & 33 $\pm$ 4     & \ \ 97 $\pm$ 1    & 79 $\pm$ 1      & 25 $\pm$ 2 & \ \ 86 $\pm$ 2    & \ 47 $\pm$ 3      & \ \ 8 $\pm$ 4  & \ \ 73 $\pm$ 3    & 53 $\pm$ 6      & \ \ 9 $\pm$ 5 & \ \ \ $58\pm0.9$  \\
(0.015, 0.03)   & \textbf{100} $\pm$ \textbf{0}  & \textbf{95} $\pm$ \textbf{1} & 51 $\pm$ 3 & \  \textbf{100} $\pm$ \textbf{0} & 91 $\pm$ 5 & 43 $\pm$ 6 & \ \ \textbf{95} $\pm$ \textbf{1}    & \ 68 $\pm$ 5 & 25 $\pm$ 6 & \ \ 75 $\pm$ 1 & 63 $\pm$ 1 & 23 $\pm$ 1 & \ \ \ $69\pm1.0$ \\
\textbf{(0.03, 0.03)}           & \textbf{100} $\pm$ \textbf{0}         & 94 $\pm$ 2          & \textbf{54} $\pm$ \textbf{3}        &  \ \textbf{100} $\pm$ \textbf{0}         & 89 $\pm$ 3          & \textbf{47} $\pm$ \textbf{5}        &  \ \ 92 $\pm$ 1         & \ \textbf{74} $\pm$ \textbf{2}          & \textbf{29} $\pm$ \textbf{3}       &  \ \ 83 $\pm$ 1         & \textbf{68} $\pm$ \textbf{7}          & \textbf{29} $\pm$ \textbf{2}        & \ \ \ \textbf{72} $\pm$ \textbf{0.9}   \\ 
(0.06, 0.03)            & \textbf{100} $\pm$ \textbf{0}  & 94 $\pm$ 2 & 49 $\pm$ 4 & \ \textbf{100} $\pm$ \textbf{0} & 94 $\pm$ 5 & 45 $\pm$ 1 & \ \ 91 $\pm$ 1 & \ 70 $\pm$ 2 & 28 $\pm$ 5 & \ \ 63 $\pm$ 1 & 52 $\pm$ 5 & 23 $\pm$ 1 & \ \ \ $67\pm0.8$ \\
(0.03, 0)            & \textbf{100} $\pm$ \textbf{0}    & 83 $\pm$ 3      & 31 $\pm$ 3    & \ \ 98 $\pm$ 2    & 79 $\pm$ 5      & 23 $\pm$ 6 & \ \ 93 $\pm$ 1    & \ 50 $\pm$ 4      & \ \ 8 $\pm$ 4      & \ \ \textbf{85} $\pm$ \textbf{1}    & 52 $\pm$ 7      & \ \ 7 $\pm$ 3     & \ \ \ $59\pm1.1$ \\
(0.03, 0.015)            & \textbf{100} $\pm$ \textbf{0}    & \textbf{95} $\pm$ \textbf{2} & \textbf{54} $\pm$ \textbf{3} & \ \textbf{100} $\pm$ \textbf{0} & 91 $\pm$ 1 & 44 $\pm$ 6 & \ \ 92 $\pm$ 1 & \ 71 $\pm$ 2 & 27 $\pm$ 3 & \ \ 82 $\pm$ 2 & 67 $\pm$ 2 & 22 $\pm$ 3 & \ \ \ $70\pm0.8$ \\
(0.03, 0.06)            & \textbf{100} $\pm$ \textbf{0}  & 94 $\pm$ 2 & \textbf{54} $\pm$ \textbf{5} & \ \textbf{100} $\pm$ \textbf{0}  & \textbf{95} $\pm$ \textbf{2} & \textbf{47} $\pm$ \textbf{6} & \ \ 90 $\pm$ 1 & \ 70 $\pm$ 1 & \textbf{29} $\pm$ \textbf{2} & \ \ 62 $\pm$ 2 & 57 $\pm$ 9 & 16 $\pm$ 4 & \ \ \ $68\pm1.1$ \\
\bottomrule
\end{tabular}}

\vspace{-6pt}
\label{lossweightsablation}
\end{table*}

\begin{table*}[t!]
\centering
\vspace{-6pt}
\caption{Ablation study on mimic loss weights ($\lambda_1^{'}, \lambda_2^{'}$) on Traffic-school. We show navigation success rate $(\%)$ in different test conditions. We bold ($\lambda_1^{'}, \lambda_2^{'}$) = (0.03, 0.03) as this set of weights performs the best and is the one we use.}
\resizebox{1 \textwidth}{!}{
\begin{tabular}{lccclccclccclccclc}
\toprule
& \multicolumn{3}{c}{\ Training} & \multicolumn{3}{c}{\ New weather} & \multicolumn{3}{c}{\ New town} & \multicolumn{3}{c}{\ New town/weather} &  \multirow{2}{*}{\ Mean}     \\
($\lambda_1^{'}, \lambda_2^{'}$)          & Empty       & Regular      & Dense      & \ \ Empty       & Regular      & Dense      & \ \ Empty       & Regular      & Dense      &  \ \ Empty       & Regular      & Dense      & \ \  \\ 
\midrule
(0, 0.03)          & $28\pm3$        & $19\pm2$         & $22\pm5$       & \ \ $21\pm1$      & $17\pm2$       & $10\pm3$         &  \ \ \ \ $7\pm1$          & \ \ $8\pm2$           & \ \ $2\pm1$ & \ \ \ \ $6\pm2$          & $11\pm3$       & \ \ $4\pm0$   & \ \ \ $13\pm0.7$            \\
(0.015, 0.03)   & $82\pm2$ & $77\pm1$ & $38\pm2$ & \ \ $75\pm4$ & $68\pm5$ & $37\pm5$ & \ \ $36\pm1$ & \  $25\pm0$ & $11\pm4$ & \ \ $14\pm2$ & $19\pm1$ & \ \ $9\pm2$ & \ \ \ $41\pm0.8$ \\
\textbf{(0.03, 0.03)}           & 90 $\pm$ 2        & 79 $\pm$ 1         & 43 $\pm$ 5       & \ \  \textbf{83} $\pm$ \textbf{3}        & \textbf{73} $\pm$ \textbf{1}         & 39 $\pm$ 4       & \ \  \textbf{46} $\pm$ \textbf{2}        & \ 39 $\pm$ 3         & 12 $\pm$ 2       & \ \  15 $\pm$ 3        & \textbf{25} $\pm$ \textbf{2}         & \textbf{14} $\pm$ \textbf{0}       & \ \ \ $\textbf{47} \pm \textbf{0.8}$ \\                  
(0.06, 0.03)            & \textbf{91} $\pm$ \textbf{1} & \textbf{80} $\pm$ \textbf{3} & $41\pm1$ & \ \ $77\pm2$ & $69\pm3$ & \textbf{41} $\pm$ \textbf{3} & \ \ $38\pm2$ & \ $35\pm2$ & $14\pm3$ & \ \ \textbf{21} $\pm$ \textbf{1} & $23\pm3$ & $11\pm1$ & \ \ \ $45\pm0.7$ \\
(0.03, 0)            & $36\pm3$        & $28\pm3$         & $16\pm3$       & \ \ $23\pm3$      & $26\pm0$       & $11\pm3$         & \ \ \ \ $6\pm1$          & \ $10\pm1$           & \ \ $2\pm1$ & \ \ \ \ $7\pm3$          & $11\pm5$       & \ \ $5\pm1$   &  \ \ \ $15\pm0.8$            \\
(0.03, 0.015)            & $87\pm1$ & \textbf{80} $\pm$ \textbf{2} & \textbf{45} $\pm$ \textbf{6} & \ \ $76\pm6$ & $71\pm4$ & $33\pm5$ & \ \ $45\pm1$ & \ \textbf{41} $\pm$ \textbf{4} & \textbf{17} $\pm$ \textbf{1} & \ \ $18\pm3$ & $23\pm5$ & $11\pm1$ & \ \ \ $46\pm1.1$ \\
(0.03, 0.06)            & $86\pm1$ & $78\pm2$ & $42\pm2$ & \ \ $73\pm2$ & $71\pm3$ & $38\pm4$ & \ \ $37\pm2$ & \ $37\pm2$ & $14\pm1$ & \ \ $12\pm2$ & $18\pm9$ & \ \ $9\pm1$ & \ \ \ $43\pm1.0$ \\
\bottomrule
\end{tabular}}

\vspace{-6pt}
\label{lossweightsablationnoviolation}
\end{table*}

\begin{table}[t]
\centering
\vspace{-6pt}
\caption{Network architecture of the squeeze network. The dimension format is $height {\times} width {\times} channel$ for feature maps or just $channel$ for feature vectors.}
\resizebox{0.8 \textwidth}{!}{

\begin{tabular}{lcccl}
\toprule
Module        & \ \ Layer & \ \ Input Dimension & \ \ Output Dimension   \\ 
\midrule
Segmentation Mask & \ \ ResNet34 & \ \ $88 \times 200 \times 6$ & \ \ 512  \\ \cline{1-4}
\multirow{2}{*}{Stop Intentions} & \ \ \multirow{2}{*}{FC} & \ \ 3   & \ \ 128     \\ 
& & \ \ 128 & \ \ 128 \\ \cline{1-4} 
\multirow{2}{*}{Self-Speed} & \ \ \multirow{2}{*}{FC} & \ \ 1   & \ \ 128    \\ 
& & \ \ 128 & \ \ 128 \\ \cline{1-4}
Joint Embedding & \ \ FC & \ \ 512 + 128 + 128 & \ \ 512 \\ \cline{1-4}
\multirow{3}{*}{Controls} & \ \ \multirow{3}{*}{FC} & \ \ 512 & \ \ 256 \\ 
& & \ \ 256 & \ \ 256 \\ 
& & \ \ 256 & \ \ 3 \\
\bottomrule

\end{tabular}
}
\vspace{-6pt}
\label{expertmodel}
\end{table}

\begin{table}[t!]
\centering
\vspace{-6pt}
\caption{Network architecture of the mimic network. The dimension format is $height {\times} width {\times} channel$ for feature maps or just $channel$ for feature vectors.}
\resizebox{0.8 \textwidth}{!}{

\begin{tabular}{lcccl}
\toprule
Module        & \ \ Layer & \ \ Input Dimension & \ \ Output Dimension   \\ 
\midrule
Seg Mask Embedding & \ \ ResNet34 & \ \ $88 \times 200 \times 3$ & \ \ 512  \\ \cline{1-4}
Stop Intentions Embedding & \ \ ResNet34 & \ \ $88 \times 200 \times 3$   & \ \ 128     \\  \cline{1-4} 
\multirow{2}{*}{Self-Speed} & \ \ \multirow{2}{*}{FC} & \ \ 1   & \ \ 128    \\ 
& & \ \ 128 & \ \ 128 \\ \cline{1-4}
Joint Embedding & \ \ FC & \ \ 512 + 128 + 128 & \ \ 512 \\ \cline{1-4}
\multirow{3}{*}{Controls} & \ \ \multirow{3}{*}{FC} & \ \ 512 & \ \ 256 \\ 
& & \ \ 256 & \ \ 256 \\ 
& & \ \ 256 & \ \ 3 \\
\bottomrule

\end{tabular}
}
\label{drivingmodel}
\vspace{-6pt}
\end{table}

\begin{table}[t]
\centering
\vspace{-6pt}
\caption{Network architecture of the stop intention estimation network used in Two-stage-(F).}
\resizebox{0.8 \textwidth}{!}{

\begin{tabular}{lcccl}
\toprule
Module        & \ \ Layer & \ \ Input Dimension & \ \ Output Dimension   \\ 
\midrule
Perception & \ \ ResNet34 & \ \ $ 88 \times 200 \times 3$ & \ \ 512  \\ \cline{1-4}
\multirow{3}{*}{Stop Intentions} & \ \ \multirow{3}{*}{FC} & \ \ 512 & \ \ 256 \\ 
& & \ \ 256 & \ \ 256 \\ 
& & \ \ 256 & \ \ 3 \\
\bottomrule

\end{tabular}
}
\vspace{-6pt}
\label{intentionsmodel}
\end{table}

\section{Detailed Model Architectures for Ablation Study}

\subsection{Our SAM Model}
Network architectures of the squeeze network and the mimic network are explained in Tab.~\ref{expertmodel} and Tab.~\ref{drivingmodel}, respectively.

\subsection{Two-stage-(F)}
The two-stage-(F) model uses ErfNet \cite{erfnet} for semantic segmentation, which is also used in \cite{Muller2018DPT}, and a ResNet34 based network, Tab.~\ref{intentionsmodel}, for stop intentions estimation. The two-stage-(F) model then feeds the estimated segmentation masks and three-category intention values to a network with the same architecture as the squeeze network. In two-stage, we directly use the pretrained weights of the squeeze network, which is trained on ground-truth semantic segmentation and stop intention values. In two-stage-F, we further fine-tune the squeeze network on estimated perception inputs to reduce the impact of propagating estimation errors of the perception inputs.

\begin{table}[t]
\centering
\vspace{-6pt}
\caption{Network architecture of the segmentation mask decoder used in Multi-task and Feature mimic. Conv/Deconv (k, s, i, o, nb) denotes a Conv/DeConv layer with (kernel size, stride, input channels, output channels, no bias), and BN denotes a batch normalization layer.}
\resizebox{1.0 \textwidth}{!}{
\begin{tabular}{lccl}
\toprule
Layer        & \ \ Input Dimension & \ \ Output Dimension \\ 
\midrule
FC & \ \ 512 & \ \ 1536 \\ \cline{1-3}
Reshape & \ \ 1536 & \ \ $1 \times 3 \times 512$ \\ \cline{1-3}
Deconv (k2, s2, i512, o512, nb), BN, ReLU & \ \ $ 1 \times 3 \times 512$ & \ \ $ 3 \times 7 \times 512$ \\
Conv (k3, s1, i512, o512, nb), BN, ReLU & \ \ $3 \times 7 \times 512$ & \ \ $3 \times 7 \times 512$ \\ \cline{1-3}
Deconv (k3, s2, i512, o256, nb), BN, ReLU & \ \ $3 \times 7 \times 512$ & \ \ $6 \times 13 \times 256$  \\
Conv (k3, s1, i256, o256, nb), BN, ReLU & \ \ $6 \times 13 \times 256$ & \ \ $6 \times 13 \times 256$ \\ \cline{1-3}
Deconv (k3, s2, i256, o128, nb), BN, ReLU & \ \ $6 \times 13 \times 256$ & \ \ $11 \times 25 \times 128$ \\
Conv (k3, s1, i128, o128, nb), BN, ReLU & \ \ $11 \times 25 \times 128$ & \ \ $11 \times 25 \times 128$  \\ \cline{1-3}
Deconv (k3, s2, i128, o64, nb), BN, ReLU & \ \ $11 \times 25 \times 128$ & \ \ $22 \times 50 \times 64$ \\
Conv (k3, s1, i64, o64, nb), BN, ReLU & \ \ $22 \times 50 \times 64$ & \ \ $22 \times 50 \times 64$ \\ \cline{1-3}
Deconv (k3, s2, i64, o64, nb), BN, ReLU & \ \ $22 \times 50 \times 64$ & \ \ $44 \times 100 \times 64$ \\
Conv (k3, s1, i64, o64, nb), BN, ReLU & \ \ $44 \times 100 \times 64$ & \ \ $44 \times 100 \times 64$ \\ \cline{1-3}
Deconv (k3, s2, i64, o64, nb), BN, ReLU & \ \ $44 \times 100 \times 64$ & \ \ $88 \times 200 \times 64$ \\
Conv (k3, s1, i64, o64, nb), BN, ReLU & \ \ $88 \times 200 \times 64$ & \ \ $88 \times 200 \times 64$ \\ \cline{1-3}
Conv (k3, s1, i64, o6, nb) & \ \ $88 \times 200 \times 64$ & \ \ $88 \times 200 \times 6$ \\

\bottomrule

\end{tabular}
}
\vspace{-6pt}
\label{segmaskdecoder}
\end{table}

\begin{table}[t!]
\centering
\vspace{-6pt}
\caption{Network architecture of the intentions decoder used in Multi-task and Feature mimic.}
\resizebox{0.6 \textwidth}{!}{

\begin{tabular}{lcccl}
\toprule
Layer & \ \ Input Dimension & \ \ Output Dimension   \\ 
\midrule
\ \ \multirow{3}{*}{FC} & \ \ 128 & \ \ 256 \\ 
& \ \ 256 & \ \ 256 \\ 
& \ \ 256 & \ \ 3 \\
\bottomrule

\end{tabular}
}
\label{intentionsdecoder}
\vspace{-6pt}
\end{table}

\subsection{Multi-task}
For the multi-task model, we use the same architecture as the mimic network except we add two additional decoders that use the outputs of the segmentation mask embedding and stop intentions embedding branches to estimate segmentation mask and stop intentions, respectively. The architecture of the segmentation mask decoder is given in Tab.~\ref{segmaskdecoder}. This decoder consists of several deconvolution (\textit{Deconv}) blocks (a deconvolution layer and a convolution layer), with skip connections between the outputs of the \textit{Deconv} block and the outputs of the corresponding ResNet block from the ResNet34 encoder. The intentions decoder (Tab~\ref{intentionsdecoder}) has a similar architecture to the \textit{Controls} module for the mimic network except that it takes in a vector of length 128, the length of the intentions embedding.

\vspace{-2pt}
\subsection{Feature mimic}
For the feature mimic model, we train two networks for semantic segmentation and stop intentions estimation. Both networks use identical encoder architectures as the corresponding ResNet branches of the mimic network; this choice aids mimicking these networks' representations. For the decoders, we use the same decoder architectures as the multi-task model (Tab.~\ref{segmaskdecoder} for segmentation mask decoder and Tab.~\ref{intentionsdecoder} for intentions decoder).

\subsection{Res101-NM}
For the Res101-NM model, the architecture is the same as that of the mimic network except the segmentation mask and stop intentions embedding branches are replaced by a single ResNet101 branch with output FC vector size of $640 = 512 + 128$ to keep total latent embedding size the same as the mimic network.

\section{Efficient Seg and Control Mimic Baselines}

We construct comparable baselines, efficient seg and control mimic, for  LEVA~\cite{behl2020label} and LBC~\cite{Chen2019lbc}, respectively. Descriptions for both baselines are provided below.

\subsection{Efficient seg}

Similar to LEVA, we construct a two-stage pipeline that uses coarse segmentation masks: we first estimate coarse segmentation masks and stop intentions and then train a driving network to drive using the estimated coarse masks and stop intentions. 

Specifically, we train three networks:
\begin{enumerate}
  \item we first train a network to estimate dense segmentation masks (the ErfNet~\cite{erfnet} model used in two-stage-(F)). We then coarsen the estimated annotations for the \textit{pedestrians}, \textit{vehicles}, and \textit{trafficSigns} classes using bounding boxes extracted from the estimated segmentation mask; LEVA coarsens the annotations for similar classes. We choose to estimate dense segmentation masks and then coarsen them as opposed to estimating coarse segmentation masks directly as we found that the first method yields better performance.
  \item a network to estimate intentions (the same intentions estimation model used in two-stage-(F))
  \item a model with the same architecture as the squeeze network trained to drive using the estimated coarse segmentation masks and stop intentions.
\end{enumerate}

\subsection{Control mimic}

Similar to LBC, we train the mimic network to mimic the privileged agent's (squeeze network's) controls; note that our SAM method instead mimics the squeeze network's embeddings. The control loss used is the same as used in our method except that we use the squeeze network's estimated controls on the training set as ground-truth controls instead. We also adopt the ``whiteboxing" data augmentation technique used in LBC: for all frames in the training set, we mimic the squeeze network's estimated controls for all four high-level turning commands $c = $ $\{{\rm follow, left, right, straight}\}$, as opposed to only the turning command associated with each frame.

\section{Training Dataset}

We collect a 10 hours ($\sim$ 360K frames from the front view, 10 FPS) training dataset using the same data collector as \cite{Codevilla_2019_ICCV}. The dataset is collected in the training town (Town01) and four training weathers (``Clear Noon", ``Heavy Rain Noon", ``Clear Noon After Rain", and ``Clear Sunset"), identical to the training conditions for all the evaluation benchmarks. For each episode, we randomly sample the number of vehicles and pedestrians from the ranges [30, 60] and [50, 100], respectively.

\section{Semantic Segmentation}
For the semantic segmentation annotations, we retain classes relevant to driving and throw out nuisance classes. Specifically, we use the \textit{pedestrians}, \textit{roads}, \textit{vehicles}, and \textit{trafficSigns} classes and map the \textit{roadlines} and \textit{sidewalks} classes to the same class. We map all other classes to a nuisance class. Hence, we obtain a total of 6 classes for the semantic segmentation.

\section{CILRS Original v.s. Rerun}

We rerun CILRS~\cite{Codevilla_2019_ICCV} using the author-provided code and model. Comparisons between the \textit{original} and the \textit{rerun} results are shown in Tab.~\ref{rerunnocrash} and Tab.~\ref{reruntrafficlight}. We notice that some of the \textit{rerun} numbers differed significantly (by more than 5\%) from those reported in the original CILRS paper. For these numbers, we report the numbers we obtained from rerunning their released code and model.

\begin{table*}[t!]
\centering
\vspace{-6pt}
\caption{Comparison on NoCrash as reported in the \textit{original} CILRS paper~\cite{Codevilla_2019_ICCV} v.s. our \textit{rerun} with author-released code and model. We show navigation success rate $(\%)$ in different test conditions. Columns with * indicate evaluation settings where we report  numbers from the \textit{rerun} since the success rate differences are larger than 5\%; otherwise, we report numbers from the \textit{original} CILRS paper.}
\resizebox{1 \textwidth}{!}{

\begin{tabular}{lccclccclccclccclc}
\toprule
                & \multicolumn{3}{c}{\ Training} & \multicolumn{3}{c}{\ New weather} & \multicolumn{3}{c}{\ New town} & \multicolumn{3}{c}{\ New town/weather} &  \multirow{2}{*}{\ Mean}     \\
Method          & Empty       & Regular      & Dense      & \ \ Empty       & Regular      & Dense*      & \ \ Empty       & Regular      & Dense      &  \ \ Empty*       & Regular      & Dense      & \ \  \\ 
\midrule
CILRS Original &  $\textbf{97}\pm\textbf{2}$         & $\textbf{83}\pm\textbf{0}$          & $42\pm2$        &  \ \ $96\pm1$         & $77\pm1$          & $\textbf{47}\pm\textbf{5}$        &  \ \ $\textbf{66}\pm\textbf{2}$         & \ $49\pm5$          & $\textbf{23}\pm\textbf{1}$        & \ \  $\textbf{90}\pm\textbf{2}$         & $56\pm2$          & $24\pm8$        & \ \ \ $\textbf{63}\pm\textbf{1.0}$  \\
CILRS Rerun             & $93 \pm 1$         & $\textbf{83} \pm \textbf{2}$          & $\textbf{43}\pm \textbf{2}$       & \ \  $\textbf{99} \pm \textbf{1}$         & $\textbf{81} \pm \textbf{2}$          & $39\pm5$        & \ \ $65 \pm 2$         & \ $\textbf{51} \pm \textbf{1}$          & $\textbf{23}\pm\textbf{1}$        &  \ \ $66 \pm 2$         & $\textbf{59} \pm \textbf{2}$         & $\textbf{27} \pm \textbf{3}$        & \ \ \ 61 $\pm$ 0.7  \\ 
\bottomrule
\end{tabular}

}
\label{rerunnocrash}
\vspace{-6pt}
\end{table*}

\begin{table}[t]
\centering
\vspace{-6pt}
\caption{Comparison on traffic light success rate (percentage of not running the {\em red} light) as reported in the \textit{original} CILRS paper \cite{Codevilla_2019_ICCV} v.s. our \textit{rerun} using author-released code and model. We note that the CILRS paper does not report standard deviations, and results on new weather as well as new town conditions. In general, the \textit{original} numbers are comparable with our \textit{rerun} results.}
\resizebox{0.8 \textwidth}{!}{

\begin{tabular}{lccccc}
\toprule
Method         & \ \ \ Training & \ \ New weather & \ \ New town & \ \ New town/weather & \ \ Mean\\ 
\midrule
CILRS Original & \ \ \ 53                            & \ \ N/A                             & \ \ N/A                             & \ \ 36                            & \ \ 45                       \\
CILRS Rerun & \ \ \ $59\pm2$                            & \ \ $32\pm1$                             & \ \ $43\pm1$                             & \ \ $35\pm2$                            & \ \ $42\pm0.8$                                              \\
\bottomrule
\end{tabular}
}
\vspace{-6pt}
\label{reruntrafficlight}
\end{table}

\section{Q\&A}

In this section, we address hypothetical questions that may arise from this work.\\

\noindent \textbf{1. Why do you provide comparable baselines for LEVA~\cite{behl2020label} and LBC~\cite{Chen2019lbc} but not LSD~\cite{ohn2020learning}?}

\noindent We provide comparable baselines for LEVA~\cite{behl2020label} and LBC~\cite{Chen2019lbc} so that we can compare our method of leveraging side tasks to their methods of leveraging side tasks on a fair basis (comparable backbones and same dataset). In contrast, for LSD~\cite{ohn2020learning}, such a comparison is moot as LSD does not focus on leveraging side tasks but instead combines reinforcement learning with a multimodal mixture of experts driving model, an orthogonal direction to our work. For the best performance, one can integrate leveraging side tasks with an ensemble of driving models and reinforcement learning, but this is out of the scope of our work.

\noindent \textbf{2. What are the advantages of Traffic-school over reporting separate metrics for different types of infractions?}

\noindent Though CARLA reports various driving infractions, these statistics are scattered and hard to analyze, leading to prior works computing different infractions metrics that are not directly comparable. For example, \citet{Codevilla_2019_ICCV} reports percentage of crossing red traffic lights while \citet{Chen2019lbc} reports number of traffic light violations per 10 km. To resolve this issue, previous benchmarks usually compute the route success rate, but they ignore several infraction types. We resolve both issues (scattered statistics and ignoring various infractions) by proposing Traffic-school, which computes route success rate while penalizing many basic infraction types (crashes, red light violations, out of lane violations) as failures.

\noindent \textbf{3. Why do the results on Traffic-school not monotonically decrease as we go from Empty to Regular to Dense? For example, in main paper Tab.~\ref{noviolation}, CILRS success rates in new town are 2\%, 7\%, 4\% in Empty, Regular, Dense, respectively.}

\noindent The results on Traffic-school do not necessarily decrease monotonically because the hand-coded non-player vehicles in the CARLA simulator stop for red lights. Hence, for agents that are better at stopping for vehicles than for red lights, stopping for vehicles may help with not violating red lights, leading to the performance on Traffic-school potentially increasing with a higher number of vehicles. This phenomenon ideally should not happen for agents that perform well in stopping for red lights, emphasizing the need for benchmarks such as Traffic-school that take red light violations into account.

\noindent \textbf{4. Why does the SAM model perform better for New town/weather than for New town on the CARLA benchmark (Supp Mat Tab.~\ref{carla})?}

\noindent Generally, it would be expected that New town/weather would be the hardest setting. However, previous papers \cite{Dosovitskiy2017Carla,Codevilla_2019_ICCV} have also observed this same trend for the CARLA benchmark. Since many risky driving infractions (e.g. running into opposite lane, collision, etc.) are not penalized in this saturated and flawed old benchmark, the improved metrics of New town/weather does not indicate that the agent is actually performing better under this setting. This unexpected behavior is not observed in our newly proposed Traffic-school benchmark thanks to more realistic evaluation protocols than the old benchmarks.
}

\end{document}